\documentclass[10pt]{article} 
\usepackage[preprint]{tmlr}

\usepackage{hyperref}
\usepackage{url}
\usepackage{amsfonts}       
\usepackage{amsmath}
\usepackage{mathtools}

\usepackage{physics}            
\usepackage{bm}

\usepackage{caption}
\usepackage{subcaption}          

\usepackage{caption}
\usepackage{subcaption}          
\usepackage[ruled]{algorithm2e}

\title{Pathwise Gradient Variance Reduction with Control Variates in Variational
  Inference}

\author{\name Kenyon Ng \email kenyon.ng@student.unimelb.edu.au \\
  \addr School of Mathematics and Statistics\\
  The University of Melbourne \AND
  \name Susan Wei \thanks{Main part of the work
    done while at the University of Melbourne.}
  \email susan.wei@monash.edu\\
  \addr Department of Econometrics and Business Statistics \\
  Monash University }

\DeclareMathOperator*{\argmin}{arg\,min}
\DeclareMathOperator*{\argmax}{arg\,max}

\DeclareMathOperator*{\diag}{diag}
\newcommand{\vect}[1]{\bm{#1}}

\newcommand{\dimz}{{\dim_z}}
\newcommand{\dimlbda}{{\dim_\lambda}}
\newcommand{\gelbo}{g}         
\newcommand{\gelbocv}{h}       
\newcommand{\rr}{r}             

\newcommand{\dt}{f}             

\newcommand{\data}{\mathcal{D}} 
\newcommand{\intgd}{\varphi}    
\newcommand{\cv}{c}             
\newcommand{\CV}{C}             
\newcommand{\trans}{T}  
\newcommand{\E}{\mathbb{E}}     
\newcommand{\Var}{\mathbb{V}}   
\newcommand{\Cov}{\mathbb{C}}   
\newcommand{\qlbda}{q}  
\newcommand{\qzero}{q_0}   
\newcommand{\nmc}{L}            
\newcommand{\ncv}{J}            

\newcommand\citeT[1]{%
  \citet{#1}}

\newcommand\citeP[1]{%
  \citep{#1}}

\def\month{09}  
\def\year{2024} 
\def\openreview{\url{https://openreview.net/forum?id=XXXX}} 

\begin{document}

\maketitle

\begin{abstract}
  Variational inference in Bayesian deep learning often involves computing the
  gradient of an expectation that lacks a closed-form solution. In these cases,
  pathwise and score-function gradient estimators are the most common
  approaches. The pathwise estimator is often favoured for its substantially
  lower variance compared to the score-function estimator, which typically
  requires variance reduction techniques. However, recent research suggests that
  even pathwise gradient estimators could benefit from variance reduction. In
  this work, we review existing control-variates-based variance reduction
  methods for pathwise gradient estimators to assess their effectiveness.
  Notably, these methods often rely on integrand approximations and are
  applicable only to simple variational families. To address this limitation, we
  propose applying zero-variance control variates to pathwise gradient
  estimators. This approach offers the advantage of requiring minimal
  assumptions about the variational distribution, other than being able to
  sample from it.
\end{abstract}

\section{Introduction}
Given an observed dataset $\data = \{x_{i}\}_{i=1}^{N}$ governed by a data
generating process that depends on latent variables $z \in \mathbb{R}^{\dimz}$
and a prior $p(z)$ of these latent variables, we are often interested in the
posterior distribution $p(z| \data) \propto p(\data|z) p(z)$. This posterior is
often known only up to a normalising constant and requires approximation.
Variational inference (VI) offers a way to approximate the posterior with a
simpler, tractable distribution from the variational family
$\mathcal Q = \{\qlbda(z; \lambda): \lambda \in \mathbb{R}^\dimlbda \}$. This is
typically done by minimising the Kullback-Leibler (KL) divergence from the
variational distribution $q(z; \lambda)$ to $p(z | \data)$, expressed as
$\E_{\qlbda(z; \lambda)} [\log{\qlbda(z; \lambda)} - \log{p(z | \data)} ]$, or
equivalently, maximising the evidence lower bound (ELBO)
\begin{equation*}
  \lambda^* = \argmax_\lambda \E_{\qlbda(z; \lambda)} [\log{p(z, \data)} - \log{\qlbda(z; \lambda)}],
\end{equation*}
with respect to the variational parameter $\lambda$. This approach is often
preferred to avoid computing the intractable normalising constant of
$p(z| \data)$.

The closed-form solution for $\lambda^*$ is generally unavailable. Stochastic VI
\citeP{hoffman2013}, which optimises with minibatch SGD, has revolutionised and
broadened the applications of VI. It necessitates computing the gradient of the
\textit{minibatch ELBO}, denoted as
$\operatorname{mELBO}(\lambda) = \E_{\qlbda(z; \lambda)} \left[ \rr(z; \lambda) \right]$,
where
\begin{equation}
  r(z; \lambda) = \frac{N}{B} \sum_{i \in \text{batch}} [\log{p(x_i | z)} ]+ \log{ \frac{p(z)}{\qlbda(z; \lambda)}}.
  \label{eq:rzl}
\end{equation}
Here $N$ and $B$ are the data and batch size respectively. One challenge in
stochastic VI is computing the gradient of $\operatorname{mELBO}$,
$\nabla_\lambda \E_{\qlbda(z; \lambda)} [\rr(z; \lambda)] $. As the gradient is
taken with respect to the parameter of $\qlbda$, we cannot simply push the
gradient operator through the expectation, making the computation non-trivial.
In the VI literature, there are two main types of gradient estimators for
computing $\nabla_\lambda \E_{\qlbda(z; \lambda)} [\rr(z; \lambda)] $: 1) the
\textbf{pathwise gradient estimator}, also known as the reparametrization trick;
and 2) the \textbf{score-function estimator}, or REINFORCE. The latter has
broader applicability but often comes with higher variance. Indeed, the
score-function estimator is almost always used in conjunction with control
variates to reduce its variance \citeP{ranganath2014,ji2021}. While variance
reduction for the pathwise gradient estimator is less common, recent work
suggests it may be beneficial \citeP{miller2017,geffner2020}. In this work, we
are primarily interested in reducing variance of the pathwise gradient
estimator.

The pathwise gradient estimator is readily applicable only to
\textit{reparametrizable} distributions $\qlbda(z; \lambda)$. These are
distributions where we can generate $z$ equivalently from a transformation
$z = T(\epsilon; \lambda)$, where
$\epsilon \in \mathbb{R}^{\dim z} \sim \qzero(\epsilon)$ and $\qzero$ is
referred to as the \textit{base distribution}, independent of $\lambda$. For
example, a Gaussian distribution $z \sim \mathcal{N}(\mu, \sigma^2 I)$ and its
corresponding transformation is $T(\epsilon; \lambda) = \mu + \sigma \epsilon$,
where $\epsilon$ follows a standard Gaussian distribution and
$\lambda = (\mu, \sigma)$. When $\qlbda$ is reparametrizable, we can push the
gradient operator inside the expectation, giving us the gradient of
$\operatorname{mELBO}$ as
\begin{equation}
  \label{eq:grad-elbo}
  g(\lambda) \coloneqq \nabla_\lambda \operatorname{mELBO}(\lambda) = \E_{\qzero(\epsilon)} \intgd(\epsilon; \lambda),
\end{equation}
where we define
$\intgd(\epsilon; \lambda) = \nabla_\lambda \left[ r(T(\epsilon;\lambda); \lambda) \right]$.
The pathwise gradient estimator is then a Monte Carlo estimator
of~\eqref{eq:grad-elbo} using samples $\{\epsilon_{[l]}\}^\nmc_{l=1}$ from
$\qzero$
\begin{equation}
  \label{eq:pathwise-est}
  \hat \gelbo(\epsilon_{[1]}, \ldots, \epsilon_{[\nmc]}; \lambda)
  \coloneqq \frac{1}{\nmc} \sum_{l=1}^\nmc \intgd(\epsilon_{[l]}; \lambda).
\end{equation}
We will refer to $\nmc$ as the number of gradient samples.

The variance of the gradient estimator
$ \Var [\hat \gelbo] = \E \lVert \hat \gelbo \rVert^2 - \lVert \E \gelbo \rVert^2 = \frac{1}{\nmc} \Var [\intgd]$
is thought to play a significant role in the convergence properties of the mELBO
optimizer. Here, the expectations and variances are taken with respect to
$\qzero$ --- from this point on, any expectations or variances without a
subscript refer to $\qzero$. To reduce the variance of \eqref{eq:pathwise-est},
we can add a \textit{control variate} (CV),
$c(\cdot) \in \mathbb{R}^{\dimlbda}$, to the pathwise gradient estimator
\begin{equation}
  \label{eq:grad-elbo-cv-single}
  \frac{1}{\nmc} \sum_{l=1}^\nmc \left[ \intgd(\epsilon_{[l]}; \lambda) + c(\epsilon_{[l]}) \right],
\end{equation}
where the CV is a random variable with zero expectation, that is,
$\E [\intgd + c] = \E \intgd$. Let $\Tr(\Cov[\intgd, c])$ denote the trace of
the covariance matrix
$\Cov[\intgd, c] = \E[(\intgd - \E \intgd) (c - \E c)^\top]$. A good CV should
exhibit a strong, negative correlation with $\intgd$, since
\begin{equation}
  \label{eq:var-estimator}
  \textstyle \Var[\nmc^{-1} \sum_{l=1}^\nmc \intgd(\epsilon_{[l]}; \lambda) + c(\epsilon_{[l]})] = \Var [\hat \gelbo] + \nmc^{-1} (\Var[c] + 2 \Tr(\Cov[\intgd, c]))
\end{equation}
Therefore, as long as $\Tr(\Cov[\intgd, c]) < 0$ and
$| \Tr(\Cov[\intgd, c]) | < \frac{1}{2} \Var[c]$, the CV-adjusted gradient
estimator \eqref{eq:grad-elbo-cv-single} will exhibit a smaller variance than
\eqref{eq:pathwise-est}. Finally, we can form a CV as a linear combination of
multiple CVs. Let
$\CV : \mathbb{R}^{\dimz} \rightarrow \mathbb{R}^{\dimlbda \times \ncv}$ be a
matrix with $J$ CVs as its columns. This leads to the CV-adjusted gradient
estimator
\begin{equation}
  \label{eq:pathwise-cv-est}
  \hat \gelbo(\epsilon_{[1]}, \ldots, \epsilon_{[\nmc]}; \lambda) \coloneqq \frac{1}{\nmc} \sum_{l=1}^\nmc \left [ \intgd(\epsilon_{[l]}; \lambda) + \CV(\epsilon_{[l]}) \beta \right ],
\end{equation}
which remains a valid CV due to the linearity of expectation operators. Here
$\beta \in \mathbb{R}^\ncv$ is a vector of coefficients corresponding to each
CV. This construction allows us to combine several weak CVs into a stronger one
by adjusting $\beta$.

For obvious reasons, applying CV is only worthwhile if its computation is
cheaper than increasing the number of samples in~\eqref{eq:pathwise-est}.
From~\eqref{eq:var-estimator}, we see that the estimator variance can be halved
either by doubling $\nmc$ or halving $\Var [\intgd + \cv]$. This poses a unique
challenge when applying CV in the low $\nmc$ regime (common in VI where $\nmc$
is often very low), as the cost of CV may outweigh the cost of increasing $\nmc$
for the same variance reduction. CVs developed for Markov Chain Monte Carlo
(MCMC) do not easily apply here because 1) they require large $\nmc$, but $\nmc$
can be as small as one in stochastic VI, and 2) MCMC variance reduction is
typically done at the very end, whereas in stochastic VI, it's needed for each
gradient update.

\subsection{Contribution} This work reviews existing CV-adjusted pathwise
gradient estimators in the context of VI, primarily examining whether employing
CV leads to faster convergence in terms of wall-clock time. We are also
motivated by the gap in VI literature on gradient variance reduction when
$\qlbda$ is reparametrizable, but its mean and covariance are not available in
closed form. A good example is normalizing flow, where $z$ is the result of
pushing a base distribution $\qzero$ through an invertible transformation
$T(\cdot; \lambda)$ parameterised by $\lambda$, i.e. $z = T(\epsilon; \lambda)$
where $\epsilon \sim \qzero$. This transformation can be arbitrarily complex and
often involves neural networks. To address this, we introduce a CV-adjusted
gradient estimator based on zero-variance control variates (ZVCV)
\citeP{assaraf1999,mira2013,oates2017}, which doesn't have this limitation.

This paper is structured as follows: Section~\ref{sec:related-work} reviews the
latest advancements in variance reduction techniques for pathwise gradient
estimators in VI. Sections~\ref{sec:selecting-beta-cv}
and~\ref{sec:control-variates} discuss methods for selecting $\beta$ and $\CV$
respectively. The novel ZVCV-based method is introduced in
Section~\ref{sec:zvcv}. Finally, experimental results are presented in
Section~\ref{sec:experiments}.

\section{Related Work}
\label{sec:related-work}

Variance reduction for the pathwise gradient estimator in VI has been explored
in \citeT{miller2017} and \citeT{geffner2020}. These works focused on designing
a single CV (i.e.~$\CV$ has only one column) with the form
$\CV = \E \tilde{\intgd} - \tilde \intgd$, where
$\tilde \intgd(\epsilon; \lambda)$ approximates $\intgd(\epsilon; \lambda)$. The
expectation $\E \tilde{\intgd}$ is intended to be theoretically tractable, but
this usually places restrictions on $\trans$ (and therefore, $\qlbda$).

For instance, \citeT{miller2017} proposed a $\tilde \intgd$ based on the
first-order Taylor expansion of $\nabla_z \log p(z, \data)$. However, this
necessitates the expensive computation of the Hessian
$\nabla^2_z \log p(z, \data)$. \citeT{geffner2020} improved upon this by using a
quadratic function to approximate $\log p(z, \data)$. Their method only requires
the first-order gradient $\nabla_z \log p(z, \data)$, and their
$\E \tilde \intgd$ has a closed-form solution as a function of the mean and
covariance of $\qlbda$. This method can be further extended to accommodate
$\qlbda$ without a closed-form mean and covariance by estimating
$\E \tilde \intgd$ empirically; see Section~\ref{sec:quad-cv} for details. In
both of these work, they focused on Gaussian $\qlbda$.

Our proposed estimator based on ZVCV shares similarities with another work from
the same group in \citeT{geffner2020}. Like \citeT{geffner2018}, we propose
combining weak CVs into a stronger one. However, our work differs in how we
construct individual CVs and the optimisation criterion for $\beta$.

\section{Selecting $\beta$ for CV-adjusted pathwise gradient estimators}
\label{sec:selecting-beta-cv}
The utility of CV depends on the choice of $\beta$ and $\CV$ in
\eqref{eq:pathwise-cv-est}. In this section, we will discuss various strategies
to pick an appropriate $\beta$ given a family of $\CV$.

\subsection{A unique set of $\beta$ for each dimension of $\lambda$}
\label{sec:full-per-dimension}
The formulation in~\eqref{eq:pathwise-cv-est} suggests that the same set of
$\beta$ is used across the dimensions of $\intgd$. This can be too restrictive
for $\CV$ that are weakly-correlated to $\intgd$. In such instance, having a
unique set of $\beta$ coefficients for each dimension of $\intgd$ can be
beneficial, as it allows the coefficients to be selected on a per-dimension
basis. In fact, this can be easily done by turning $\CV$ into a
$\dimlbda \times (\ncv \dimlbda)$-dimensional, block diagonal matrix
$\diag(\CV_{1, :}, \ldots, \CV_{\dimlbda, :})$, where $\CV_{i, :}$ is the
$i^{th}$ row of the original $\CV$. In other words, we expand the number of CV
to $\ncv \dimlbda$, and each CV will only reduce the variance of a single
dimension of $\intgd$.

\subsection{Optimisation criteria for $\beta$}
\label{sec:optim-crit-beta}
The $\beta$ is usually chosen to minimise the variance of~$\hat \gelbocv$. In
practice, this variance is evaluated empirically due to the lack of its
closed-form expression. There are three approaches in the literature, the first
of which is a direct approximation of the variance with samples
$\{\epsilon_{[l]}\}^\nmc_{l=1}$,
\begin{equation*}
  \textstyle \Var[\intgd + \CV \beta] \approx
  \frac{1}{\nmc(\nmc-1)} \sum_{l>l'} \lVert \intgd(\epsilon_{[l]})
  + \CV(\epsilon_{[l]}) \beta
  - \intgd(\epsilon_{[l']})
  - \CV(\epsilon_{[l']}) \beta \rVert^2,
\end{equation*}
as seen in \citeT{belomestny2018}. The second approach is based on the
definition of variance
\begin{equation}
  \label{eq:ls-representation}
  \textstyle \Var[\intgd + \CV \beta] = \E \lVert \intgd - \E[\intgd + \CV \beta] + \CV \beta \rVert^2
  \approx \underset{\alpha \in \mathbb{R}^\dimlbda}{\min} \frac{1}{\nmc}
  \sum^\nmc_{l=1} \lVert \intgd(\epsilon_{[l]})
  + \alpha + C(\epsilon_{[l]}) \beta \rVert^2,
\end{equation}
where $\alpha \in \mathbb{R}^\dimlbda$ is an intercept term in place of the
unknown $\E[\intgd + \CV \beta]$. This is cheaper to compute than the former as
it only requires $O(\nmc)$ operations rather than $O(\nmc^2)$ \citeP{si2022}.
Finally, the third approach relies on the assumption that $\E[\CV \beta] = 0$
and is based on the observation that
$\Var[\intgd + \CV \beta] = \E \lVert \intgd + \CV \beta \rVert^2 - \lVert \E \intgd \rVert^2$.
This suggests that $\Var[\intgd + \CV \beta]$ can be equivalently minimised by
solving $\beta^* = \argmin_\beta \E \lVert \intgd + \CV \beta \rVert^2$, the
solution of which is given
\begin{equation}
  \label{eq:esn-general}
  \beta^* = - \E [\CV^\top \CV]^{-1} \E [\CV^\top \intgd].
\end{equation}
See \citeT{geffner2018} for the derivation. This approach, however, often
requires estimating the expectations empirically and performing a costly
inversion of size $J$ matrix.

In the development of our CV, we focus on the second approach
~\eqref{eq:ls-representation} as this is generally the cheapest among the three.
This approach is also equivalent to solving a linear least squares problem
\begin{equation}
  \begin{bmatrix}
    \label{eq:ols-residual}
    \alpha^* \\
    \beta^*
  \end{bmatrix}
  = \argmin_{\alpha, \beta}
  \sum^\nmc_{l=1} \left\lVert
  \intgd(\epsilon_{[l]}) +
  \begin{bmatrix}
    \mathbb{I}_\dimlbda & C(\epsilon_{[l]})
  \end{bmatrix}
  \begin{bmatrix}
    \alpha \\
    \beta
  \end{bmatrix}
  \right\rVert^2,
\end{equation}
and has a unique and closed-form solution when the corresponding design matrix
is full-column rank. Even when such condition is not satisfied, a penalty term
can be added to the objective function of \eqref{eq:ols-residual} and we end up
with a penalised least squares problem \citeP{south2022}. Alternatively, we can
solve~\eqref{eq:ols-residual} with an iterative optimisation algorithm to obtain
a (non-unique) solution \citeT{si2022}.

\section{Control variates}
\label{sec:control-variates}
Having reviewed methods to select $\beta$ given a family of $\CV$, we now turn
our attention to constructing $\CV$. We will first propose a simple modification
of \citeT{geffner2020} that will enable it to work for variational distributions
$\qlbda$ with unknown mean and covariance. Subsequently, we will introduce ZVCV,
which can be constructed without the knowledge of $\qlbda$ or $\trans$.

\subsection{Quadratic approximation control variates}
\label{sec:quad-cv}
In this section, we review the quadratic-approximation CV proposed in
\citeT{geffner2020}. An important distinction at the outset is their assumption
that the entropy term in $\operatorname{mELBO}$,
$-\E_{\qlbda(z)} \log \qlbda(z; \lambda)$, is \textit{known}. As such the focus
of \citeT{geffner2020} is to reduce the variance of
$\E \nabla_\lambda f(\trans(\epsilon;\lambda))$, where
\begin{equation}
  f(z) = \frac{N}{B} \sum_{i \in \text{batch}} [\log{p(x_i | z)} ]+ \log p(z).
  \label{eq:fzl}
\end{equation}
The CV is based on the quadratic approximation of \eqref{eq:fzl},
$\tilde{\dt}(z ; v) = b^\top_v (z - z_0) + \frac{1}{2} (z - z_0)^\top B_v (z - z_0)$,
and has the form of
\begin{equation}
  \label{eq:quad-cv}
  \CV(\epsilon)
  = \E[ \nabla_\lambda \tilde \dt (\trans(\epsilon; \lambda)) ]
  - \nabla_\lambda \tilde \dt (\trans(\epsilon; \lambda)).
\end{equation}
Here, $v = \{B_v, b_v\}$ are the parameters of the quadratic equation that are
chosen to minimise the $L^2$ difference between $\nabla \dt(z)$ and
$\nabla \tilde \dt(z)$. We will drop $v$ in $\tilde \dt$ for the sake of
brevity. The location parameter $z_0$ is set to $\E \trans(\epsilon; \lambda)$.
This quadratic approximation of $\dt$ can also be viewed as a linear
approximation on $\nabla \dt$.

The first term in~\eqref{eq:quad-cv} has a closed-form expression that depends
on the mean and covariance of $\qlbda(z; \lambda)$, making the expectation cheap
to evaluate when they are readily available. However, this is not the case when
$\trans(\epsilon; \lambda)$ is arbitrarily complex, e.g.~normalizing flow. A
direct workaround of this limitation is to replace
$ \E \nabla_\lambda \tilde \dt (\trans(\epsilon; \lambda)) $ with its empirical
estimate based on samples of $\epsilon$. Note that $\tilde \dt$ requires
$z_0 = \E \trans(\epsilon; \lambda)$, which we estimate using another
independent set of $\epsilon$. See Algorithm~\ref{alg:quad} for a summary of the
procedure.

As~\eqref{eq:fzl} is a part of the Monte Carlo
estimator~\eqref{eq:pathwise-cv-est}, it could be tempting to estimate
$\E \nabla_\lambda \tilde \dt (\trans(\epsilon; \lambda))$ with an average of
the $\nabla_\lambda \tilde \dt (\trans(\epsilon; \lambda))$ evaluations that
have been computed in~\eqref{eq:pathwise-cv-est}. This is to be avoided as it
will result in the two terms in~\eqref{eq:quad-cv} cancelling each other out.

As~\eqref{eq:quad-cv} is designed to be strongly correlated with $\intgd$ when
$\tilde \dt$ is reasonably close to $\dt$, the choice of $\beta$ becomes less
significant. \citeT{geffner2020} opted to minimise the expected squared norm
with a scalar $\beta$ (note that $\CV$ is a column vector in this case), the
solution of which is given in~\eqref{eq:esn-general}. In their work, the
expectations $\E[\CV^\top \intgd]$ and $\E[\CV^T \CV]$ are replaced with their
empirical estimates computed from $\CV$ and $\intgd$
in~\eqref{eq:pathwise-cv-est}, instead of fresh evaluations. However, the
resulting gradient estimate is biased due to the dependency of between $\beta$
and $\CV$, as $\E [\CV (\epsilon) \beta(\epsilon)] \neq 0$ in general.

While this bias is not mentioned explicitly in \citeT{geffner2020}, we
conjecture that they overcame the issue by estimating the expectations with
$\CV$ and $\intgd$ computed from previous iterations, as specified in
Algorithm~\ref{alg:quad}. Therefore, their $\beta$ is independent from $\CV$ in
the current iteration. This will avoid introducing bias to the gradient estimate
at the cost of having sub-optimal $\beta$. They also claimed that their
estimates of $\E[\CV^\top \intgd]$ and $\E[\CV^T \CV]$ (and by extension,
$\beta$) do not differ much across iterations. Moreover, their $\beta$ is
largely acting as an auxiliary `switch' of the CV when $\tilde \dt$ is a poor
approximation of $\dt$, rather than the primary mechanism to reduce the
estimator variance, since the $\beta$ will be almost 0 when $\tilde \dt$ is not
approximating well (i.e.~$\Cov[\intgd, \CV] \approx 0$). Their $\CV$ only kicks
in when it is sufficiently correlated to $\intgd$.

Lastly, we return to the discussion of the entropy term at the beginning of this
section. Our setup is more general than~\citeT{geffner2020} as we does not
assume the entropy term~$-\E_{\qlbda(z)} \log \qlbda(z; \lambda)$ to necessarily
have a closed-form expression, i.e.~our $r(z, \lambda)$ includes
$-\log q(z; \lambda)$. Although it was claimed in~\citeT{geffner2020} that their
quadratic approximation CV can also be similarly designed for $\rr(z,\lambda)$
in \eqref{eq:rzl} rather than $f(z,\lambda)$ in \eqref{eq:fzl}, we found the
implementation difficult because the updating step of $v$ requires the gradient
$\nabla_z \log \qlbda(z; \lambda)$, and in turn $\pdv{\lambda}{z}$, which is
challenging to compute.

\subsection{Zero-variance control variates}
\label{sec:zvcv}

The CV in \citeT{geffner2020} require one to know the mean and covariance of
$\qlbda(z; \lambda)$. To avoid this requirement, we propose the use of
gradient-based CV \citeP{assaraf1999,mira2013,oates2017}. These CV are generated
by applying a Stein operator $\mathcal{L}$ to a class of user-specified
functions $P(z)$. Typically the Stein operator uses $\nabla_z \log q(z)$, the
gradients of the log probability density function for the distribution over
which the expectation is taken, but it does not require any other information
about $\intgd$ or $\trans$.

We will focus on the form of gradient-based CV known as ZVCV
\citeP{assaraf1999,mira2013}. ZVCV uses the second order Langevin Stein operator
and a polynomial $P(z)=\sum_{j=1}^\ncv \beta_j P_j(z)$, where $P_j(z)$ is the
$j$th monomial in the polynomial and $\ncv$ is the number of monomials. The CV
are
\begin{align*}
  \{\mathcal{L} P_j(z)\}_{j=1}^\ncv = \{\Delta_z P_j(z) + \nabla_z P_j(z) \cdot \nabla_z \log q(z)\}_{j=1}^\ncv,
\end{align*}
where $\Delta_z$ is the Laplacian operator and $q(z)$ is the probability density
function for the distribution over which the expectation is taken. A sufficient
condition for these CV to have zero expectation is that the tails of $q$ decay
faster than a polynomial rate \cite[Appendix B]{oates2016}, which is satisfied
by Gaussian $q$ for example.

In this paper, we only consider first-order polynomials, so there are
$\ncv=\dimz$ number of CV of the form
$\left \{\pdv{}{z_j} \log q(z) \right\}_{j=1}^{\dimz}$. Here, $z_j$
refers to the $j^{th}$ dimension of $z$. We do not find second-order polynomials
to have any advantage over first-order polynomials; see
Appendix~\ref{sec:zvcv-hyperparameters} for a discussion. For pathwise gradient
estimators using a standard Gaussian as the base distribution, these CV simplify
further to $\{- \epsilon_j \}^{\dimz}_{j=1}$. We are also using the same set of
CV across different dimensions of $\intgd$, but assigning each dimension with a
unique set of $\beta$. That is, the matrix $\CV$ is a block-diagonal matrix
$\CV(\epsilon) = \diag(-\epsilon^{\top}, \ldots, -\epsilon^{\top})$ of size
$\dimlbda \times \dimlbda \dimz$. This is in contrast to~\citeT{geffner2020}
where the values in $\CV$ is different across dimensions, but their $\beta$ is
shared. The simplicity of ZVCV comes with the drawback that it is often not as
correlated as the integrand. This makes the choice of $\beta$ crucial.

As discussed in Section~\ref{sec:selecting-beta-cv}, the unique, closed-form
solution of~\eqref{eq:ols-residual} requires the corresponding design matrix to
have a full column rank. In our application where the models often has $\dimz$
much greater than $\nmc$, this is not the case as the corresponding design
matrix is very wide. While this problem can be solved by adding a penalty term
in \eqref{eq:ols-residual} to shrink $\beta$ towards 0
\citeP{geffner2018,south2022}, solving penalised least squares remains
prohibitively expensive as it still involves inverting a matrix of size $\dimz$.
Instead, we propose mimicking penalised least squares by
minimising~\eqref{eq:ols-residual} with respect to $\alpha$ and $\beta$ with
gradient descent. This is done by
\begin{enumerate}
  \item Initialise $\alpha$ at $-\frac{1}{\nmc} \sum^\nmc_{l=1} \intgd(\epsilon_{[l]})$
        and $\beta$ at the zero vector. Set $\gamma^{(\alpha, \beta)}$ to a low value;
  \item Take a descent step
        $(\alpha_{m+1}, \beta_{m+1}) \leftarrow (\alpha_{m}, \beta_{m}) - \gamma^{(\alpha,
            \beta)} \frac{1}{\nmc} \sum^\nmc_{l=1} \nabla_{\alpha, \beta} \lVert
          \intgd(\epsilon_{[l]}) + \alpha_m + C(\epsilon_{[l]}) \beta_m \rVert^2$;
        \label{item:inner-gd-2}
  \item Repeat Step~\ref{item:inner-gd-2} for a few times.
\end{enumerate}
See Algorithm~\ref{alg:zvcv} for a complete description. The combination of learning
rate and number of iterations is analogous to the penalty in penalised least squares: a
lower number of iterations and learning rate $\gamma^{(\alpha, \beta)}$ will result in a
near-zero $\beta$ that results from a stronger penalty (more shrinkage of $\beta$
towards 0). This procedure is also similar in spirit to \citeT{si2022}.

\section{Experiments}
\label{sec:experiments}
In these experiments, we assess the efficacy of various gradient estimators by
performing VI on the following model-dataset pairs: logistic regression on the
\textit{a1a} dataset, a hierarchical Poisson model on the \textit{frisk}
dataset, and a Bayesian neural network (BNN) on a subset of the \textit{redwine}
dataset. For the BNN model, we consider both full-batch and mini-batch (size 32)
gradient estimators. We utilise diagonal and low-rank Gaussian distributions, as
well as Real NVP \citeP{dinh2017} as our variational family $\qlbda$.
Additionally, we vary the number of gradient samples, setting $\nmc = 10$ and
$50$, and compare three types of estimators: the vanilla estimator without any
CV (NoCV), ZVCV-GD as described in Section~\ref{sec:zvcv}, and QuadCV proposed
by \citeT{geffner2020}. We report the ELBO, wall-clock time, and variance of the
gradient estimators. Comprehensive setup details are provided in
Appendix~\ref{sec:experiment-setup}.

\subsection{ELBO against wall-clock time} 
\begin{figure}[h!]
  \centering
  \begin{subfigure}[t]{\linewidth}
    \centering
    \includegraphics[width=\linewidth]{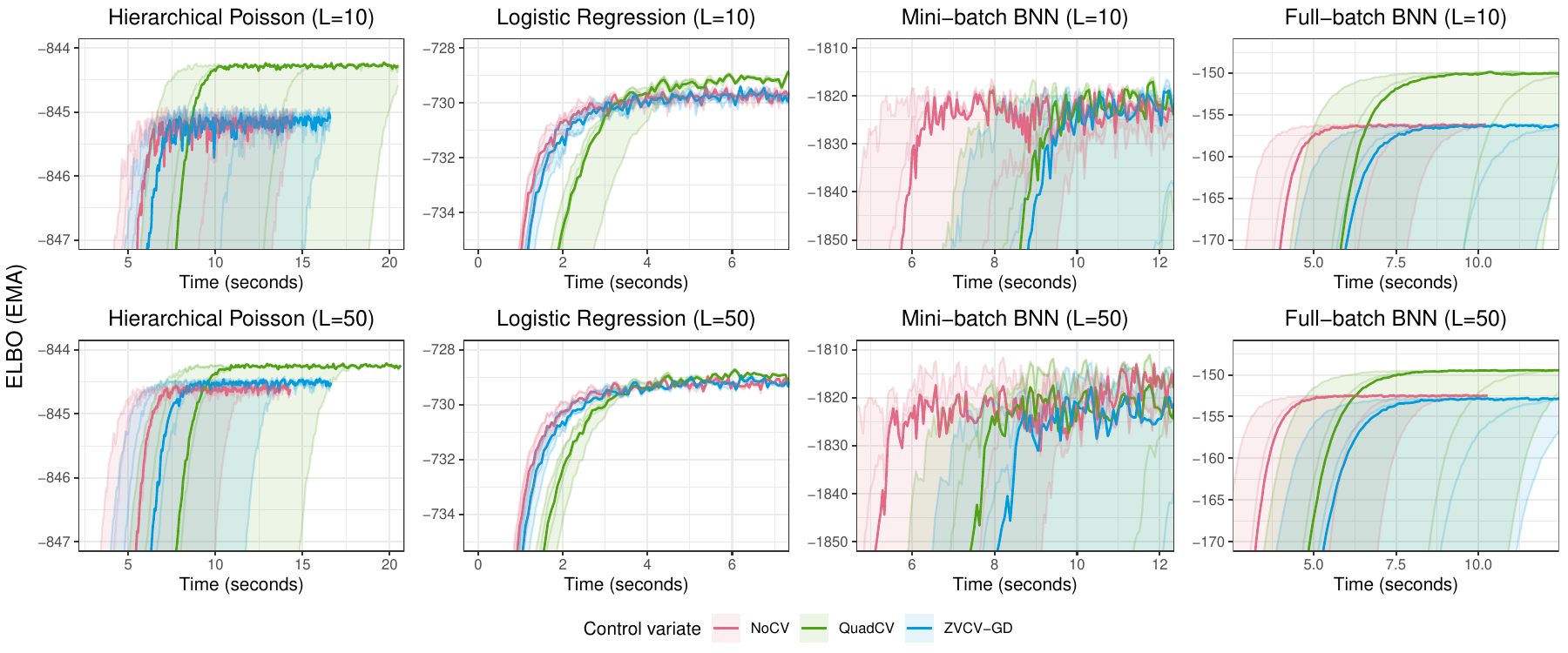}
    \caption{Mean-field Gaussian with 10 (top) and 50 (bottom) gradient samples.}
    \label{fig:elbo-time-diag}
  \end{subfigure} \\
  \vspace{5mm}
  \begin{subfigure}[t]{\linewidth}
    \centering
    \includegraphics[width=\linewidth]{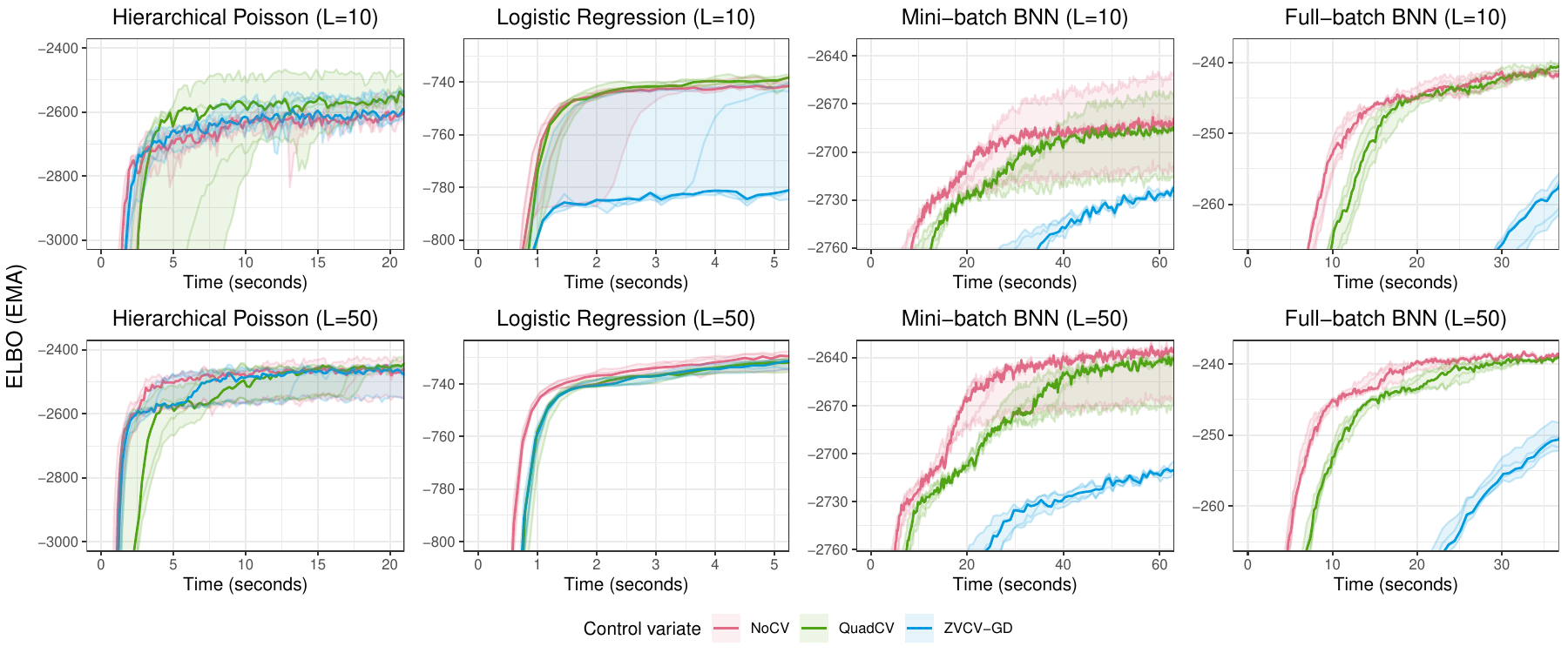}
    \caption{Real NVP with 10 (top) and 50 (bottom) gradient samples.}
    \label{fig:elbo-time-nvp}
  \end{subfigure}
  \caption{ELBO is plotted against wall-clock time for different numbers of gradient
    samples $\nmc$ and two families of $\qlbda$. The bold lines represent the median of
    ELBO values recorded at the same iteration across five repetitions. The shaded area
    illustrates the range of ELBO values across five repetitions. The ELBO values are
    smoothed using an exponential moving average. A higher ELBO indicates better
    performance. See Figure~\ref{fig:mean-elbo-time} for plots where the bold lines
    represent the mean ELBO.}
  \label{fig:elbo-time}
\end{figure}

To assess whether the computational
expense of calculating CV or additional gradient samples justifies the potential
improvement in ELBO, we measure ELBO against wall-clock time, as illustrated in
Figure~\ref{fig:elbo-time}. Our experiments reveal that NoCV generally converges
to a respectable ELBO more swiftly. Furthermore, the performance gap between the
estimators is even narrower when $\nmc=50$. An unexpected observation is that
increasing $\nmc$ from 10 to 50 incurs negligible computational cost but produce
meaningfully faster convergence, as evident when comparing the top and bottom
rows of Figure~\ref{fig:elbo-time-diag} and~\ref{fig:elbo-time-nvp}. It is
important to note that the computational cost of extra gradient samples may vary
depending on the construction of $\intgd$, and increasing $\nmc$ might not
always be a worthwhile strategy for achieving faster convergence (see, for
example, the BNNs experiments in Figure~\ref{fig:elbo-time-diaglr} of
Appendix~\ref{sec:low-rank-gaussian}).

QuadCV does succeed in increasing the maximum achievable ELBO in certain
scenarios, albeit at the expense of longer convergence times. For instance,
QuadCV can improve ELBO by approximately 0.7 nats and 6 nats in hierarchical
Poisson and full-batch BNN when using a mean-field Gaussian $\qlbda$ at
$\nmc=10$. However, this comes at a cost of roughly 50\% to 100\% more runtime
compared to NoCV. Given finite computational resources and the absence of a
universal guarantee that a slight ELBO increase will substantially enhance
downstream metrics
\citeP{yao2018a,yao2019,foong2020,masegosa2020,deshpande2022}, it is left to
practitioners to determine whether implementing CV is a worthwhile
endeavour.

\subsection{Additional experiments}
In addition to our main results, we present further findings in the Appendix to
explore the efficacy of CV under various settings. Specifically, Appendix
\ref{sec:median-elbo-iteration} examines ELBO and variance reduction against
iteration count, Appendix~\ref{sec:low-rank-gaussian} explores low-rank Gaussian
as $\qlbda$, Appendix~\ref{sec:mean-elbo-vr} presents mean curves of ELBO,
Appendix~\ref{sec:different-initialisation} provides an individual analysis of
each curve, and Appendix~\ref{sec:zvcv-hyperparameters} investigates different
variants of ZVCV. Overall, we find that while CV can reduce variance in the
gradient estimator, the computational overhead does not justify its
implementation compared to simply increasing the number of gradient samples.

\section{Conclusion}
In our study of the pathwise gradient estimator in VI, we reviewed the
state-of-the-art CV for reducing gradient variance, namely the QuadCV in
\citeT{geffner2020}. We identified a gap in the literature regarding variance
reduction of pathwise gradient estimators in stochastic VI when the variational
distribution has intractable mean and covariance, making QuadCV not directly
applicable. To address this, we proposed using ZVCV, which does not assume
specific conditions on the variational distribution.

However, our empirical results showed that neither the ZVCV-adjusted nor the
QuadCV-adjusted estimator provided substantial improvement in our evaluation
metrics to justify their implementation. Instead, we found that simply
increasing the number of gradient samples was highly effective for improving
convergence time.

Stepping back, it is worth discussing the fundamental value of variance
reduction for pathwise gradient estimators in stochastic VI. Interestingly, a
dramatic reduction in gradient variance may not lead to any noticeable effect on
the ELBO. This was observed in our experiments --- even with a substantially
lower variance, the CV-adjusted gradient estimator did not meaningfully improve
the ELBO optimization objective compared to the vanilla gradient estimator. We
can thus expect that downstream metrics, such as log predictive density, will
also reveal the general ineffectiveness of equipping the gradient estimator with
a CV. These findings seem to indicate a negative phenomenon for pathwise
gradients in stochastic VI: reducing gradient variance alone is insufficient to
improve downstream performance.

In future work, we hope to explore ZVCV-adjusted gradient estimators in
generative models where they may excel. ZVCV is particularly powerful when the
distribution of interest is difficult to sample from, such as in energy-based
models \citeP{song2021}. Additionally, implicit VI methods require the
variational distribution to be reparametrizable but not the pathwise score,
$\nabla_z \log \qlbda(z; \lambda)$, to be known (e.g.~as in normalizing flows).
\citeT{titsias2019a} showed that the pathwise score can be written as an
expectation,
$ \nabla_z \log \qlbda(z; \lambda) = E_{\qlbda(\epsilon | z; \lambda)} \nabla_z \log \qlbda(z|\epsilon; \lambda)$,
where $\qlbda(\epsilon | z; \lambda) $ is the reverse conditional. In
\citeT{titsias2019a}, the expectation with respect to the reverse conditional is
based on MCMC samples. We could potentially improve efficiency by employing ZVCV
here.

\section{Acknowledgment}
The authors would like to thank Leah South for her valuable discussions on the
project. KN was supported by the Australian Government Research Training Program
Scholarship and the Fred Knight Scholarship. SW was supported by the ARC
Discovery Early Career Researcher Fellowship (DE200101253).

\bibliographystyle{tmlr}
\bibliography{paper}

\newpage

\appendix

\section{Implementation details}
We present the implementation details and pseudocode in this section.

\subsection{Potential bias in the gradient estimator}
The unbiasedness of the CV-adjusted Monte Carlo
estimator~\eqref{eq:pathwise-cv-est} relies on the assumption that the $\beta$
are independent of $\CV$, since $\E [\CV (\epsilon) \beta(\epsilon)] \neq 0$ in
general. This necessitates that $\beta$ and $\CV$ should be estimated with
independent sets of $\epsilon$ samples. However, in practice, the $\beta$ is
estimated with the same set of $\epsilon$ in $\CV$ to save computational time at
the cost of introducing bias in the gradient estimates.

\begin{algorithm}
  \caption{Quadratic approximation control variates with empirical estimates of $\E \tilde \dt$}
  \label{alg:quad}
  \KwIn{Learning rates $\gamma^{(\lambda)}$, $\gamma^{(v)}$}
  Initialise $\lambda$, $v$ and control variate weight $\beta = 0$\;
  \For{$k = 0, 1, 2, \cdots$}{

  Sample $\epsilon_{[1]}, \ldots \epsilon_{[\nmc]} \sim \qzero$ to compute
  $\intgd(\epsilon_{[l]}; \lambda_k)$\;

  Generate an independent set of 100 $\epsilon$ samples to estimate
  $z_0 = \E \trans(\epsilon; \lambda)$\;

  Generate an independent set of 100 $\epsilon$ samples to estimate
  $\E \nabla_\lambda \tilde \dt (\trans(\epsilon; \lambda); v_k)$\;

  Compute
  $\gelbocv = \frac{1}{\nmc} \sum^\nmc_{l = 1} \left [\intgd(\epsilon_{[l]}; \lambda_k) + \CV(\epsilon_{[l]}) \beta \right ]$. See~\eqref{eq:quad-cv}\;

  Take an ascent step
  $\lambda_{k+1} \leftarrow \lambda_k + \gamma^{(\lambda)} \gelbocv$\;

  Estimate $\E[\CV^\top \CV]$ and $\E[\CV^\top \intgd]$ with
  $\epsilon_{[1]}, \ldots \epsilon_{[\nmc]}$, and update $\beta$ with
  \eqref{eq:esn-general}\;

  Take a descent step
  $v_{k+1} \leftarrow v_k - \gamma^{(v)} \frac{1}{2\nmc} \sum^L_{l=1} \nabla_v \lVert \nabla_z \dt(\trans(\epsilon_{[l]}; \lambda_k)) - \nabla_z \tilde \dt(\trans(\epsilon_{[l]}; \lambda_k); v_k) \rVert^2 $\;
  }
\end{algorithm}

\begin{algorithm}
  \caption{ZVCV-GD}
  \label{alg:zvcv}
  \KwIn{Learning rates $\gamma^{(\lambda)}$, $\gamma^{(\alpha, \beta)}$}

  Initialise $\lambda$\;

  \For{$k = 0, 1, 2, \cdots$}{

    Sample $\epsilon_{[1]}, \ldots \epsilon_{[\nmc]} \sim \qzero$\;

    Compute $\intgd(\epsilon_{[l]}; \lambda_k), \forall l = 1, \ldots, \nmc$.
    See~\eqref{eq:grad-elbo}\;

    Initialise
    $\alpha_0 = -\frac{1}{\nmc} \sum^\nmc_{l=1} \intgd(\epsilon_{[l]}; \lambda_k)$
    and $\beta_0$ at the zero vector\;

    \For{$m = 0, 1, 2, \cdots$}{

      Take a descent step
      $(\alpha_{m+1}, \beta_{m+1}) \leftarrow (\alpha_{m}, \beta_{m}) - \gamma^{(\alpha, \beta)} \frac{1}{\nmc} \sum^\nmc_{l=1} \nabla_{\alpha, \beta} \lVert \intgd(\epsilon_{[l]}; \lambda_k) + \alpha_m + \CV(\epsilon_{[l]}) \beta_m \rVert^2$

    }

    Set $\beta^*$ as the final value of $\beta$ from the previous inner loop\;

    Compute
    $\gelbocv = \frac{1}{\nmc} \sum^\nmc_{l = 1} \intgd(\epsilon_{[l]}; \lambda_k) + \CV(\epsilon_{[l]}) \beta^*$\;

    Take an ascent step
    $\lambda_{k+1} \leftarrow \lambda_k + \gamma^{(\lambda)} \gelbocv$\;

  }
\end{algorithm}

\section{Experiment Setup}
\label{sec:experiment-setup}

\subsection{Models and datasets}
We perform VI on the following model-dataset pairs: logistic regression on the
\textit{a1a} dataset, a hierarchical Poisson model on the \textit{frisk}
dataset, and Bayesian neural network (BNN) on the \textit{redwine} dataset. For
the BNN model, we consider a full-batch gradient estimator trained on a subset
of 100 data points of the \textit{redwine} dataset following the experimental
setup in \cite{geffner2020} and \cite{miller2017}. We also consider a mini-batch
estimator of size 32 but trained on the full \textit{redwine} datasets. With the
exception of mini-batch BNN, these models appeared in either \citeT{geffner2020}
or \citeT{miller2017}.

\paragraph{Logistic regression with the \textit{a1a} dataset}
We extracted the $a1a$ dataset from the
\href{https://github.com/tomsons22/ABVRR/blob/main/src/datasets/a1a}{repository} hosting
\citeT{geffner2020}. We used the full dataset $\{\bm{x}_i, y_i\}^{1605}_{i=1}$ and 90\%
of the dataset for training. The response $y_i$ is binary and is modelled as
\begin{align*}
  w_0, \bm{w}           & \sim \mathcal{N}(0, 10^2)               \\
  p(y_i | \bm{x}_i , z) & = \text{Bernoulli} \left( \dfrac{1}{1 +
    \exp(-w_0 - \bm{w}^T \bm{x}_i)} \right),
\end{align*}
where~$z = \{w_0, \vect{w}\}$ and $\dimz = 120$. The size of training and test sets are
1440 and 165 respectively.

\paragraph{Hierarchical Poisson regression with the \textit{frisk} dataset}
This example is coming from \citeT{gelman2007}. We only used a subset of data
(weapon-related crime, precincts with 10\%-40\% of black proportion), as in
\citeT{miller2017} and \citeT{geffner2020}. The response $y_{ep}$ denotes the number of
frisk events due to weapons crimes within an ethnicity group $e$ in precinct $p$ over a
15-months period in New York City:
\begin{align*}
  \mu                                   & \sim \mathcal{N}(0, 10^2)                \\
  \log \sigma_\alpha, \log \sigma_\beta & \sim \mathcal{N}(0, 10^2)                \\
  \alpha_e                              & \sim \mathcal{N}(0, \sigma^2_\alpha)     \\
  \beta_p                               & \sim \mathcal{N}(0, \sigma^2_\beta)      \\
  \log \lambda_{ep}                     & = \mu + \alpha_e + \beta_p + \log N_{ep} \\
  p(y_{ep} | z)                         & = \text{Poisson} ( \lambda_{ep} ),
\end{align*}
where~$z = \{\alpha_1, \alpha_2, \beta_1, \ldots, \beta_{32}, \mu, \log \sigma_\alpha,
  \log \sigma_\beta \}$ and $\dimz = 37$. $N_{ep}$ is the (scaled) total number of arrests
of ethnicity group $e$ in precinct $p$ over the same period of time. We do not split out
a test set due to its small size (total data size is 96).

\paragraph{Bayesian neural network with the \textit{redwine} dataset}
We push a vector input $\bm{x}_i$ through a 50-unit hidden layer and ReLU
activation's to predict wine quality. The response $y_i$ is an integer from 1 to
10 (inclusive) measuring the score of red wine. We place an uniform improper
prior on the log-variance of the weights and error
\cite[Section~5.7]{gelman2013a}:
\begin{align*}
   & p(\log \alpha^2)  \propto 1, \quad \text{equivalent to $p(\alpha) \propto \alpha^{-1}$} \\
   & p(\log \tau^2)  \propto 1, \quad \text{equivalent to $p(\tau) \propto \tau^{-1}$}       \\
   & w_i  \sim \mathcal{N}(0, \alpha^2), \quad i = 1, \ldots, 651                            \\
   & y_i | \bm{x}_i, \bm{w}, \tau  \sim \mathcal{N}(\phi(\bm{x}, \bm{w}), \tau^2)
\end{align*}
where~$\phi$ is a multi-layer perception. Here,
$z = \{\log \alpha^2, \log \tau^2, \bm{w} \}$ and $\dimz = 653$. For full-batch gradient
descent, we use two mutually exclusive subsets of 100 data point as train and test sets,
as in \citeT{miller2017} and \citeT{geffner2020}. For mini-batch gradient descent, we
use 90\% of the full dataset for training and the rest for testing (size of train and
test sets are 1431 and 168 respectively).

\subsection{Variational families} Three classes of variational families are considered:
\begin{itemize}
  \item \textbf{Mean-field Gaussian} The covariance of the Gaussian distribution
        $\mathcal{N}(\mu, \Sigma)$ is parameterised by log-scale parameters,
        i.e.~$\Sigma = \diag \left( \exp ( 2\log \sigma_1, \ldots, 2\log \sigma_\dimz )
          \right)$.
  \item \textbf{Rank-5 Gaussian} The covariance of the Gaussian distribution
        $\mathcal{N}(\mu, \Sigma)$ is parameterised by a factor
        $F \in \mathbb{R}^{\dimz \cross 5}$ and diagonal components,
        i.e.~$\Sigma = F F^\top + \diag \left( \exp ( 2\log \sigma_1, \ldots, 2\log
            \sigma_\dimz ) \right)$.
  \item \textbf{Real NVP} We use a real NVP normalizing flow \citeP{dinh2017} with two
        coupling layers and compose the layers in alternate pattern. The flow has a standard
        multivariate Gaussian as its base distribution. The scale and translation networks
        have the same architecture of $8 \times 16 \times 16$ hidden units with ReLU
        activations, followed by a fully connected layer. There is an additional tanh
        activation at the tail of the scale network to prevent the exponential term from
        blowing up.
\end{itemize}
We only present the results for mean-field Gaussian and real~NVP in the main
section. The results for rank-5 Gaussian are included in
Appendix~\ref{sec:low-rank-gaussian}, as they are largely similar to those obtained for
mean-field Gaussian.

\subsection{Optimiser and learning rate}
We use an Adam optimiser and set its learning rate $\gamma^{(\lambda)} = 0.01$,
except for the BNN models with real NVP where we set
$\gamma^{(\lambda)} = 0.001$. These learning rates have been selected as the
most best options, in terms of convergence time to a respectable ELBO, from the
set of \{0.1, 0.01, 0.001, 0.0001\}.

\subsection{Initialisations} We repeated the experiment five times, each time using
different initialisations of $\lambda$ to assess the convergence performance of each
method under varying initial conditions. For the mean-field Gaussian, the $\lambda$
values were randomly sampled from a zero-mean Gaussian distribution with a scale
parameter of 0.5. In contrast, for real NVP, we initialised the $\lambda$ values using a
Glorot normal initialiser \citeP{glorot2010}. These choices of initialisers were made
deliberately to ensure a wide range of initial values, covering both favourable and
unfavourable starting points. Consequently, we expect to observe a diverse range of ELBO
trajectories.

\subsection{Control variates}
The gradient estimator is equipped with the following control variate
strategies:
\begin{itemize}
  \item \textbf{NoCV} The vanilla gradient estimator without any control variates.
  \item \textbf{ZVCV-GD} A ZVCV with $\beta$ minimising least squares with an
        inner gradient descent, as described in Algorithm~\ref{alg:zvcv} and
        Section~\ref{sec:zvcv}. We set the learning rate
        $\gamma^{(\alpha, \beta)} = 0.001$ and iterated the inner gradient
        descent 4 times for each outer Adam step. These hyperparameter choices
        may not always yield the maximum variance reduction in every situation,
        but they represent a good compromise with computation time.
        Additionally, we have discovered that prolonging the inner gradient
        descent iterations does not necessarily lead to better variance
        reduction. For a more comprehensive discussion, please refer to
        Appendix~\ref{sec:zvcv-hyperparameters}.
  \item \textbf{QuadCV} This is the original algorithm presented in~\citeT{geffner2020}
        when $\qlbda$ is Gaussian (i.e.~the mean and covariance of $\qlbda$ are readily
        available). When $\qlbda$ is real~NVP, we use Algorithm~\ref{alg:quad} and 100 samples
        to estimate $\E \trans(\epsilon; \lambda)$ and
        $\E \nabla_\lambda \tilde \dt(\trans(\epsilon; \lambda))$. The learning rate
        $\gamma^{(v)}$ is set to $\gamma^{(\lambda)}$, following the original work.
\end{itemize}
Note that above we only compare our method in detail with \citeT{geffner2020} as it is a direct
improvement of \citeT{miller2017}.

\subsection{Evaluation settings}

\paragraph{ELBO} The ELBO for evaluation purpose is always computed with the full
dataset (even when using mini-batched ELBO for optimisation) and 500 samples
from $\qlbda$.

\paragraph{Wall-clock time}
We timed our VI implementation in JAX and ran on an Nvidia A100 80GB GPU. It is
worth noting that recorded times may vary among computing platforms and
implementations, given that our code was compiled with XLA (resulting in
platform-dependent binaries) and ran without memory constraints.

\paragraph{Variance ratio} We also present the variance ratio,
$\Var[\hat \gelbocv] / \Var[\hat \gelbo]$ where $\hat \gelbo$ and
$\hat \gelbocv$ as defined in~\eqref{eq:pathwise-est}
and~\eqref{eq:pathwise-cv-est} respectively, in every 50 iterations; a ratio
less than 1 indicates a reduction in variance relative to the corresponding NoCV
with the same number of $\nmc$. The variance of the gradient estimators is
computed by repeatedly sampling 100 gradients (say,
$\hat \gelbo_{[1]}, \ldots, \hat \gelbo_{[100]}$) from the estimator and
computed with
$\Var[\hat \gelbo] \approx \frac{1}{100} \sum_j \lVert \hat \gelbo_{[j]} - ( \frac{1}{100} \sum_i \hat \gelbo_{[i]}) \rVert^2 $.

\paragraph{Computation of variance ratio}
The variance ratio $\Var [\hat \gelbocv] / \Var [\hat \gelbo]$ was computed with the
following step:
\begin{enumerate}
  \item Collect 100 samples of $\hat \gelbo$ resulting in $\{\hat \gelbo_{[j]}\}^{100}_{j=1}$;
  \item For each $\hat \gelbo_{[j]}$, compute its corresponding control-variate-adjusted gradient
        estimate $\hat \gelbocv$~\eqref{eq:pathwise-cv-est} to collect
        $\{\hat \gelbocv_{[j]}\}^{100}_{j=1}$;
  \item Estimate
        $\Var[\hat \gelbo] \approx \frac{1}{100} \sum_j \lVert \hat \gelbo_{[j]} - ( \frac{1}{100}
          \sum_i \hat \gelbo_{[i]}) \rVert^2 $. Repeat the same step for $\Var[\hat \gelbocv]$;
  \item Calculate the ratio $\Var[\hat \gelbocv] / \Var[\hat \gelbo]$.
\end{enumerate}

This ratio is designed to evaluate the effectiveness of control variates in reducing
variance relative to a corresponding gradient estimator without control
variates. Therefore, in our work, the ratio is always computed with a pair of
$\hat \gelbo$ and $\hat \gelbocv$ with the same number of samples.

\section{Median ELBO against iteration counts}
\label{sec:median-elbo-iteration}
\begin{figure}
  \centering
  \begin{subfigure}[t]{\linewidth}
    \centering
    \includegraphics[width=\linewidth]{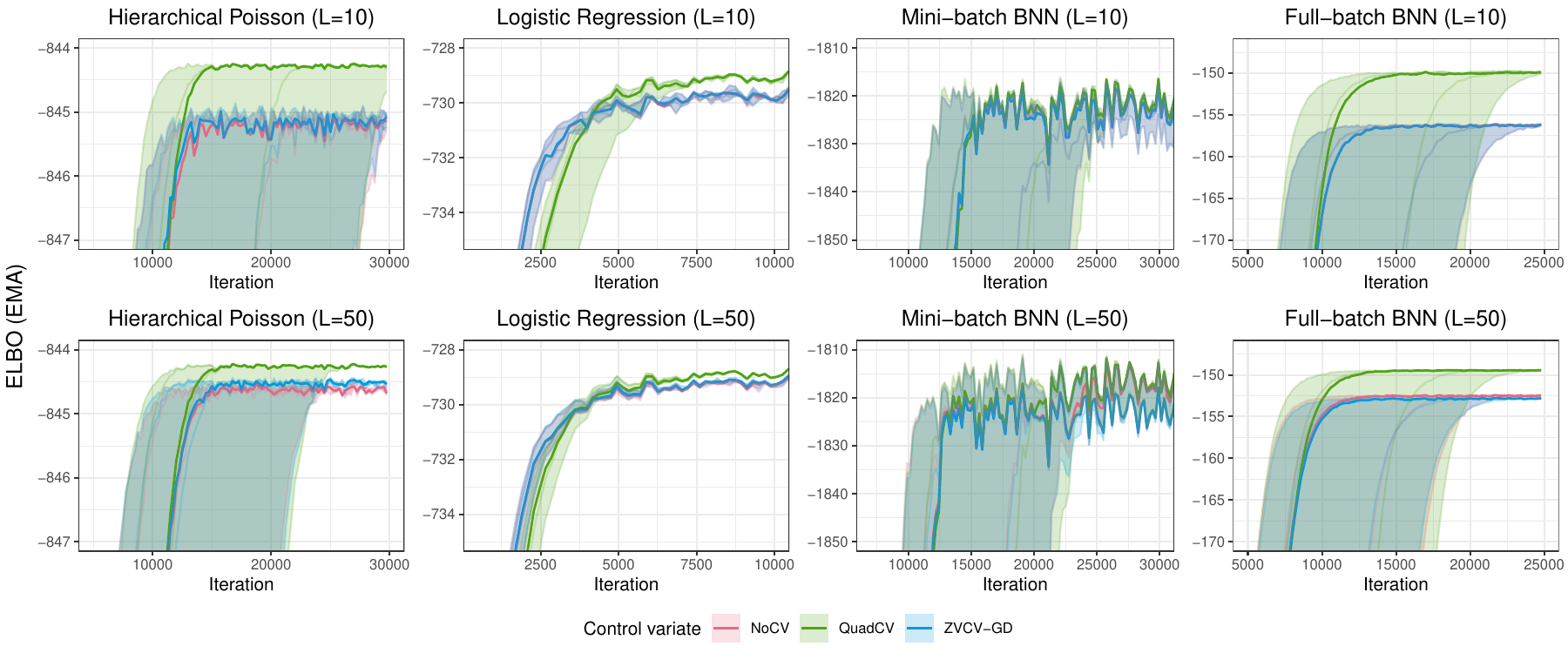}
    \caption{Mean-field Gaussian with 10 (top) and 50 (bottom) gradient samples.}
    \label{fig:elbo-iter-diag}
  \end{subfigure} \\
  \vspace{5mm}
  \begin{subfigure}[t]{\linewidth}
    \centering
    \includegraphics[width=\linewidth]{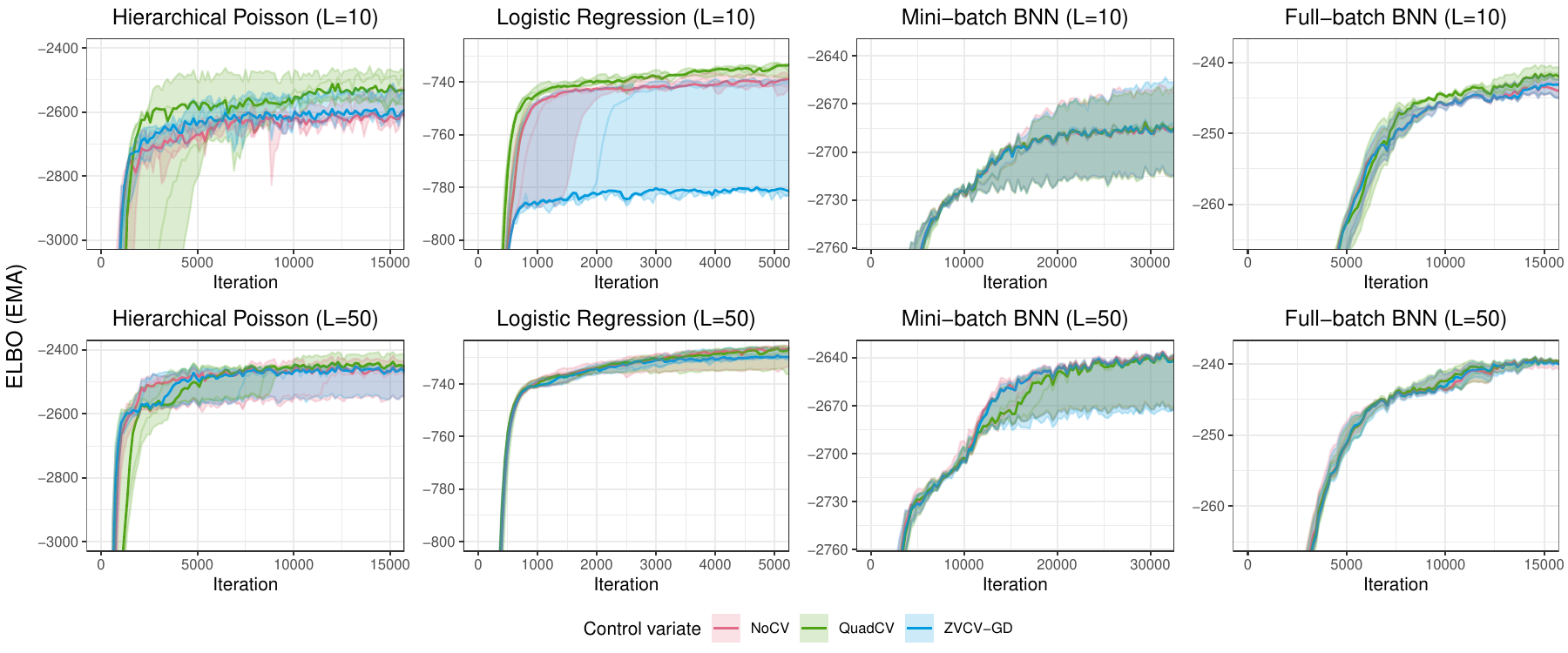}
    \caption{Real NVP with 10 (top) and 50 (bottom) gradient samples.}
    \label{fig:elbo-iter-nvp}
  \end{subfigure}
  \caption{ELBO is plotted against the number of gradient descent steps for different
    numbers of gradient samples $\nmc$ and two families of $\qlbda$. The bold lines
    represent the median of ELBO values recorded at the same iteration across five
    repetitions. The shaded area illustrates the range of ELBO values across five
    repetitions. The ELBO values are smoothed using an exponential moving average. The
    trajectories of ZVCV-GD and NoCV are nearly identical in both full-batch and
    mini-batch BNN when $L=10$. A higher ELBO indicates better performance. See
    Figure~\ref{fig:mean-elbo-iter} for plots where the bold lines represent the mean
    ELBO.}
  \label{fig:elbo-iter}
\end{figure}

The results in Figure~\ref{fig:elbo-iter} demonstrate that QuadCV generally outperform
NoCV, while ZVCV-GD provides only marginal improvement and can even converge to a
suboptimal maximum in some cases (e.g.~logistic regression, real~NVP 2, and $\nmc$ =
10). The performance gap between the estimators also decreases as the number of gradient
samples $\nmc$ increases, as seen in the bottom rows of Figure~\ref{fig:elbo-iter-diag}
and~\ref{fig:elbo-iter-nvp}. It should be noted that QuadCVs may perform poorly in the
early stages of gradient descent (e.g.~logistic regression on mean-field Gaussian and
hierarchical Poisson on real~NVP) as it takes time to learn the quadratic function
$\tilde \dt$. In general, there is also a high degree of variability in ELBO across
different runs. This is especially noticeable in Figure~\ref{fig:elbo-iter-diag} due to
the substantial impact of $\lambda$ initialisation on optimisation convergence. For a
more detailed examination of the individual trajectories with various initialisations,
please refer to Appendix~\ref{sec:different-initialisation}.

\begin{figure}
  \centering
  \includegraphics[width=\linewidth]{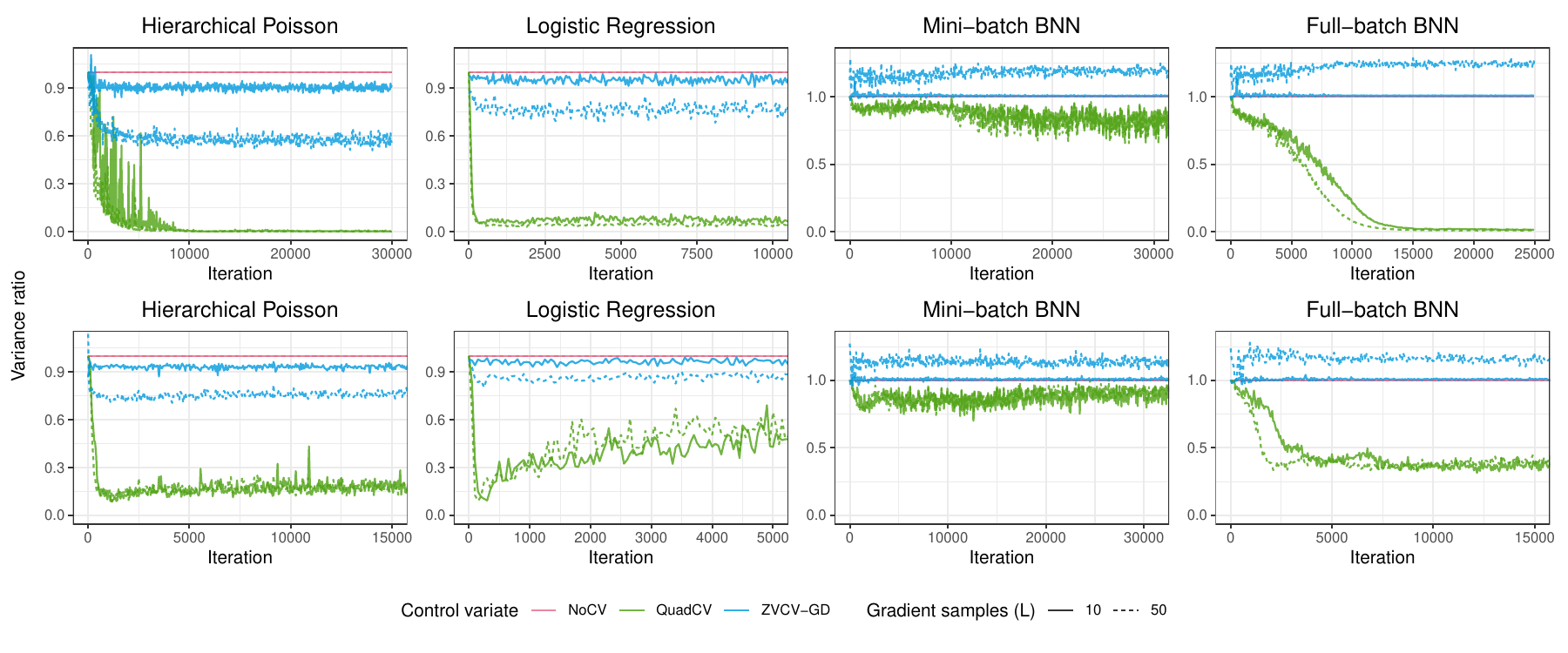}
  \caption{We present the variance ratio
    $\Var[\hat \gelbocv] / \Var[\hat \gelbo]$, where $\hat \gelbo$ is NoCV and
    $\hat \gelbocv$ is either ZVCV-GD or QuadCV, at each iteration. We show only
    the median variance ratios recorded at the same iteration across five
    repetitions, omitting the individual variance ratios from each repetition to
    prevent clutter in the plots. The ratios from mean-field Gaussian and real
    NVP are shown in top and bottom rows respectively. Note that NoCV (in red)
    is always 1 by definition. We see that ZVCV-GD (in blue) struggles to reduce
    variance in the BNN models. There is also a significant overlap in QuadCV
    between $\nmc = 10$ (solid green) and $\nmc = 50$ (dotted green). A lower
    ratio indicates better performance. See Figure~\ref{fig:mean-vr} for plots
    where the bold lines represent the mean variance ratios.}
  \label{fig:vr}
\end{figure}

The variance ratio of the gradient estimators can help explain the performance
gap observed in Figure~\ref{fig:elbo-iter}. As shown in Figure~\ref{fig:vr},
QuadCV generally achieves a lower variance than ZVCV-GD, particularly for
Gaussian $\qlbda$ when $\E \tilde \dt$ can be computed exactly. The estimator
with ZVCV-GD and larger $\nmc$ tends to perform better in models with fewer CV
(i.e.~low $\dimz$), as the $\beta$ is less susceptible to overfitting when
solving the least squares with the gradient descent algorithm discussed in
Section~\ref{sec:zvcv}. On the contrary, in models with large $\dimz$, such
as BNNs, ZVCV-GD fails to reduce variance.

A noteworthy characteristic of QuadCV is that variance reduction only becomes prominent
after $\tilde \dt$ in~\eqref{eq:quad-cv} has been adequately trained. This typically
occurs as the optimisation process nears convergence. With a QuadCV-adjusted gradient
estimator, it is possible to push the ELBO at convergence a few nats further, although
significant time has to be spent to reach convergence at all. However, this raises an
interesting question about the worthiness of such an effort, as a relatively minor
improvement in ELBO may not necessarily translate into substantially improved downstream
metrics; see Appendix~\ref{sec:different-initialisation} for a more in-depth discussion.

The comparison between $\nmc=10$ and $\nmc=50$ in Figure~\ref{fig:elbo-iter} suggests
that variance reduction in the early stages can facilitate quicker convergence in terms
of iteration counts (notice the leftward shift in the trajectories for $\nmc=50$. This
observation implies that employing a larger number of gradient samples is an effective
strategy to improve the convergence performance of stochastic VI, as long as the
computation of additional gradient samples remains cost-effective in the overall
optimisation process. It is important to note that increasing $\nmc$ from~$10$ to~$50$
immediately reduces the gradient estimator's variance by five-fold (equivalent to a
variance ratio of~0.2) from the very first iteration of the optimisation, in contrast to
QuadCV. These results suggest that variance reduction is more beneficial during the
initial stages of optimisation when the goal is to expedite convergence towards a
satisfactory ELBO, rather than aiming to attain the maximum achievable ELBO.

\section{Results from rank-5 Gaussian}
\label{sec:low-rank-gaussian}

\begin{figure}
  \centering
  \begin{subfigure}[t]{\linewidth}
    \centering
    \includegraphics[width=\linewidth]{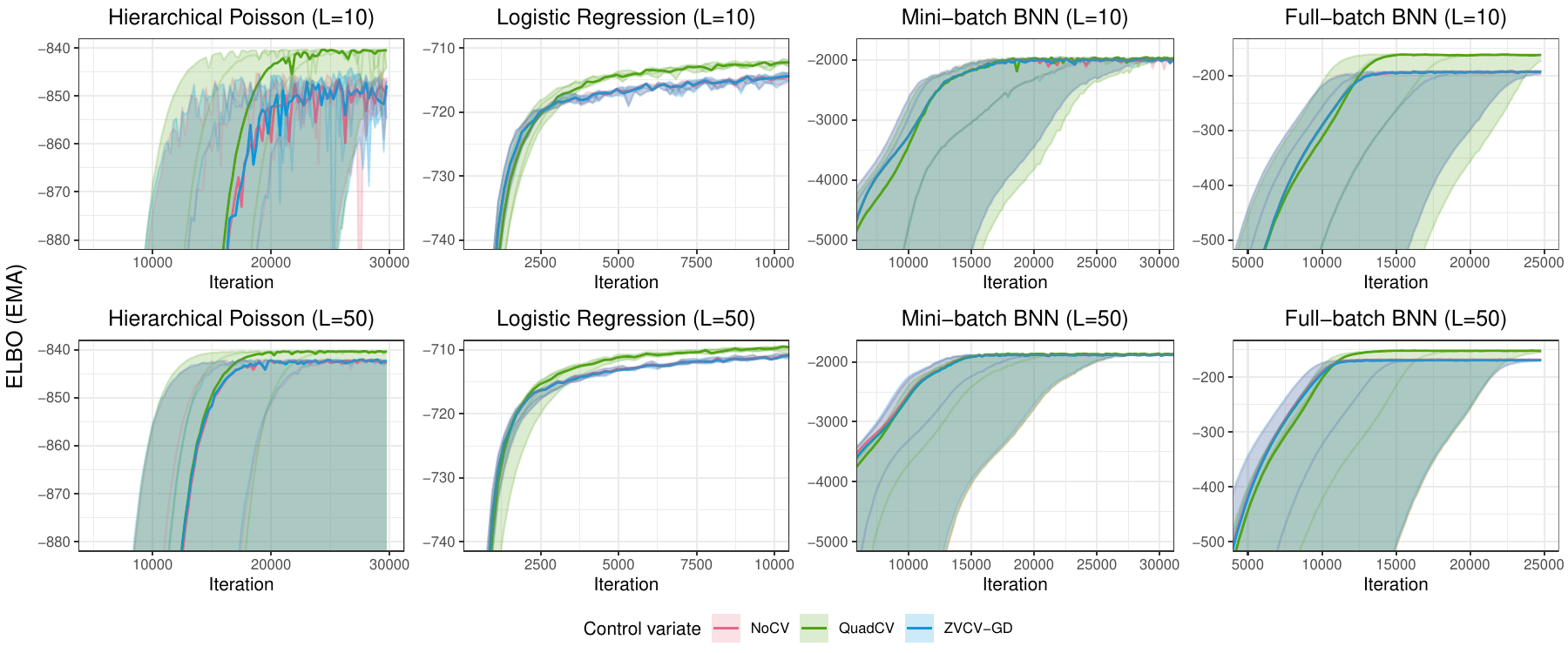}
    \caption{ELBO versus iteration counts.}
    \label{fig:elbo-iter-diaglr}
  \end{subfigure} \\
  \vspace{5mm}
  \begin{subfigure}[t]{\linewidth}
    \centering
    \includegraphics[width=\linewidth]{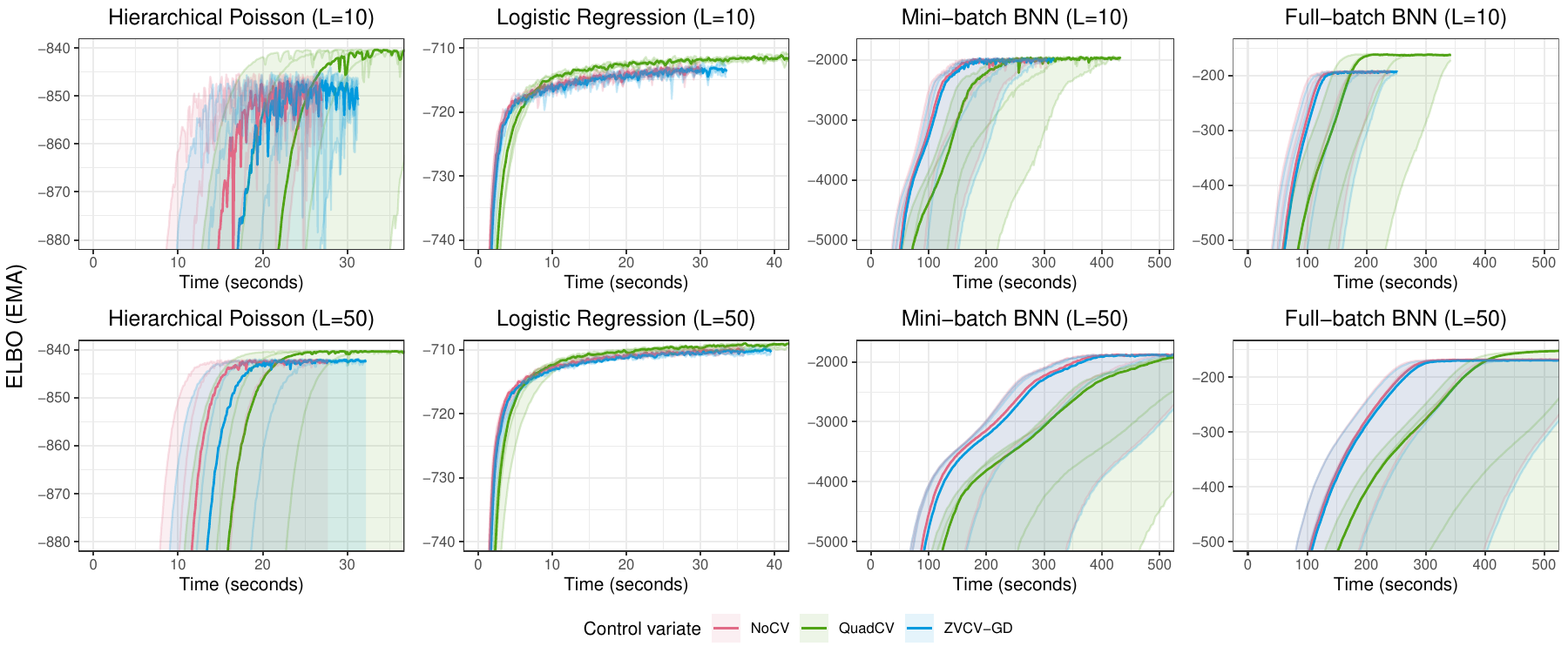}
    \caption{ELBO versus wall-clock time.}
    \label{fig:elbo-time-diaglr}
  \end{subfigure}
  \caption{ ELBO is plotted against gradient descent steps and wall-clock time for
    varying numbers of gradient samples $\nmc$ using rank-5 Gaussian. The bold lines
    represent the median of ELBO values recorded at the same iteration across five
    repetitions. The ELBO values have been smoothed using an exponential moving
    average. A higher ELBO indicates better performance. See
    Figure~\ref{fig:mean-elbo-diaglr} for plots where the bold lines represent the mean
    ELBO. }
  \label{fig:elbo-diaglr}
\end{figure}

\begin{figure}
  \centering
  \begin{subfigure}[t]{\linewidth}
    \centering
    \includegraphics[width=\linewidth]{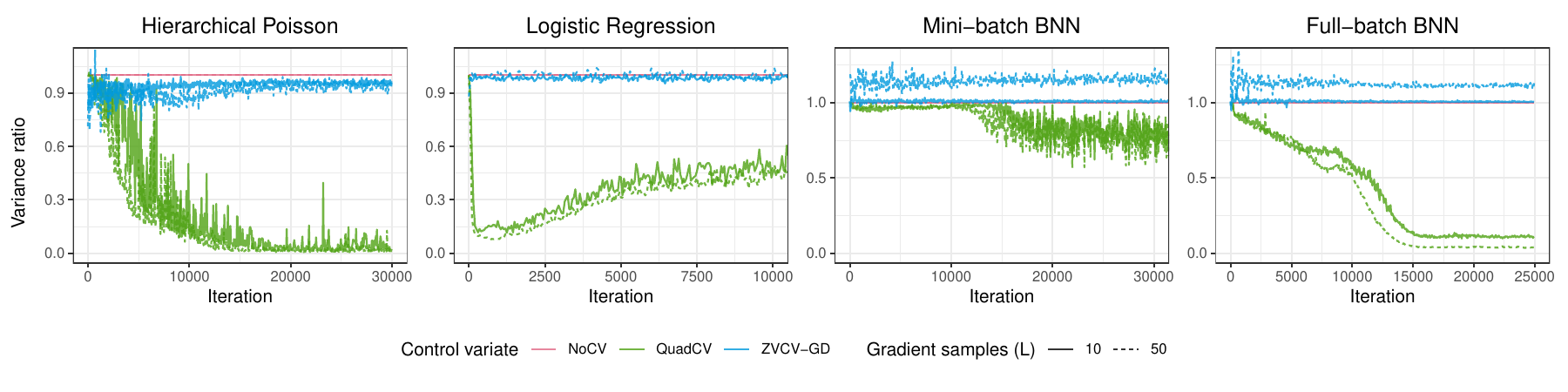}
    \caption{Variance ratio}
  \end{subfigure}
  \caption{We present the variance ratio $\Var[\hat \gelbocv] / \Var[\hat \gelbo]$ of
    rank-5 Gaussian, where $\hat \gelbo$ is NoCV and $\hat \gelbocv$ is either ZVCV-GD
    or QuadCV, at each iteration. We show only the median variance ratios recorded at
    the same iteration across five repetitions, omitting the individual variance ratios
    from each repetition to prevent clutter in the plots. Note that NoCV (in red) is
    always 1 by definition. We see that ZVCV-GD (in blue) struggles to reduce variance
    in the BNN models. There is also some overlap between $\nmc = 10$ (solid green) and
    $\nmc = 50$ (dotted green). A lower ratio indicates better performance. A lower
    ratio indicates better performance. See Figure~\ref{fig:mean-vr-diaglr} for plots
    where the bold lines represent the mean variance ratios.  }
  \label{fig:vr-diaglr}
\end{figure}

The insight derived from Figure~\ref{fig:elbo-diaglr} and~\ref{fig:vr-diaglr} below are
similar to those obtained from Figure~\ref{fig:elbo-iter},~\ref{fig:elbo-time}
and~\ref{fig:vr}. In most cases, the cost for evaluating control variates outweighs the
improvement in ELBO achieved through variance reduction in the gradient estimator. We
observe marginal gain in ELBO despite the estimators with control variates taking longer
time to converge.

\section{Mean ELBO trajectories and variance ratio}
\label{sec:mean-elbo-vr}
We have recreated the figures in Section~\ref{sec:experiments} and
Appendix~\ref{sec:low-rank-gaussian}, with the exception that the bold lines now
represent the means of ELBO or variance ratios, as opposed to their medians. Using means
provides a more transparent depiction of the robustness of each method, although it can
be substantially influenced by the repetition that starts farthest from the optimal
$\lambda$. Ideally, individual trajectories should be plotted separately (as in
Appendix~\ref{sec:different-initialisation}), but this is not feasible due to space
limitations. Nonetheless, the findings of this study are substantiated by interpreting
either the mean or median of the evaluation statistics.

\begin{figure}[h!]
  \centering
  \begin{subfigure}[t]{\linewidth}
    \centering
    \includegraphics[width=\linewidth]{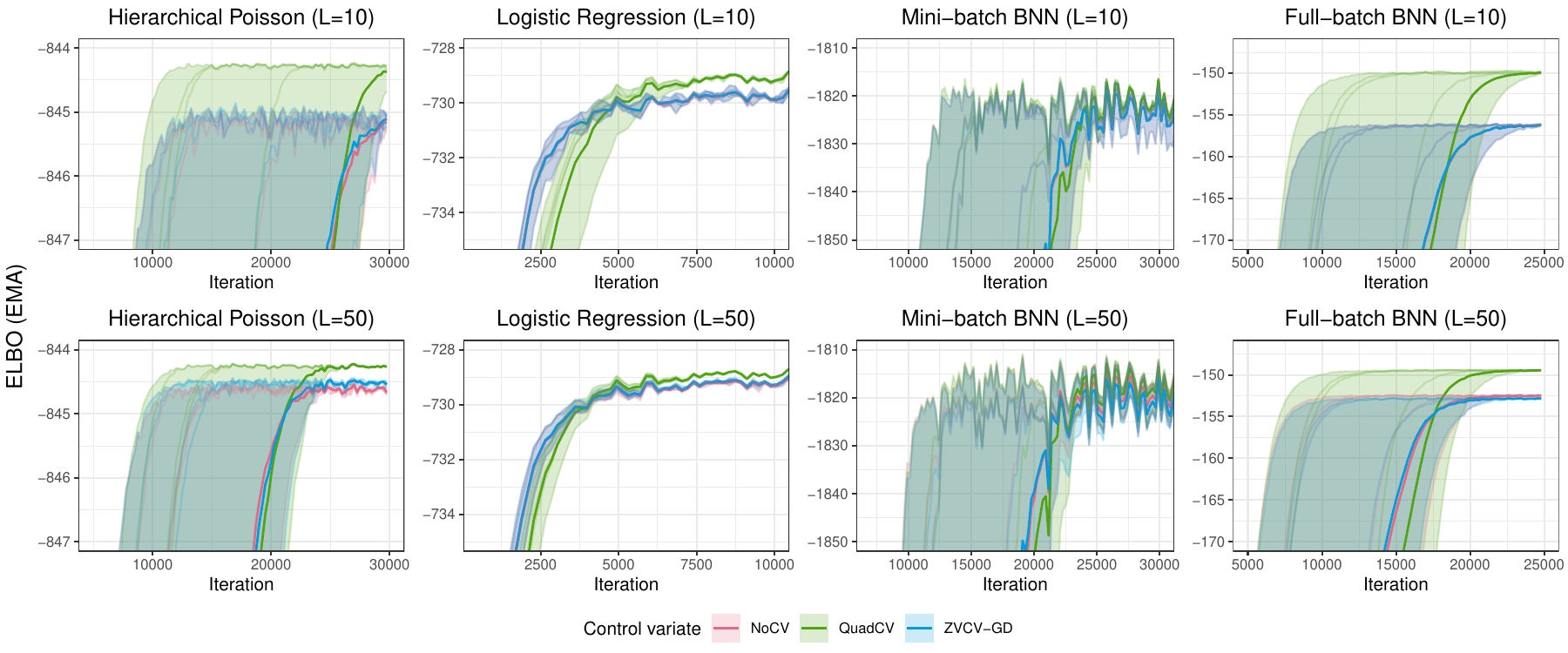}
    \caption{Mean-field Gaussian with 10 (top) and 50 (bottom) gradient samples.}
    \label{fig:mean-elbo-iter-diag}
  \end{subfigure} \\
  \vspace{5mm}
  \begin{subfigure}[t]{\linewidth}
    \centering
    \includegraphics[width=\linewidth]{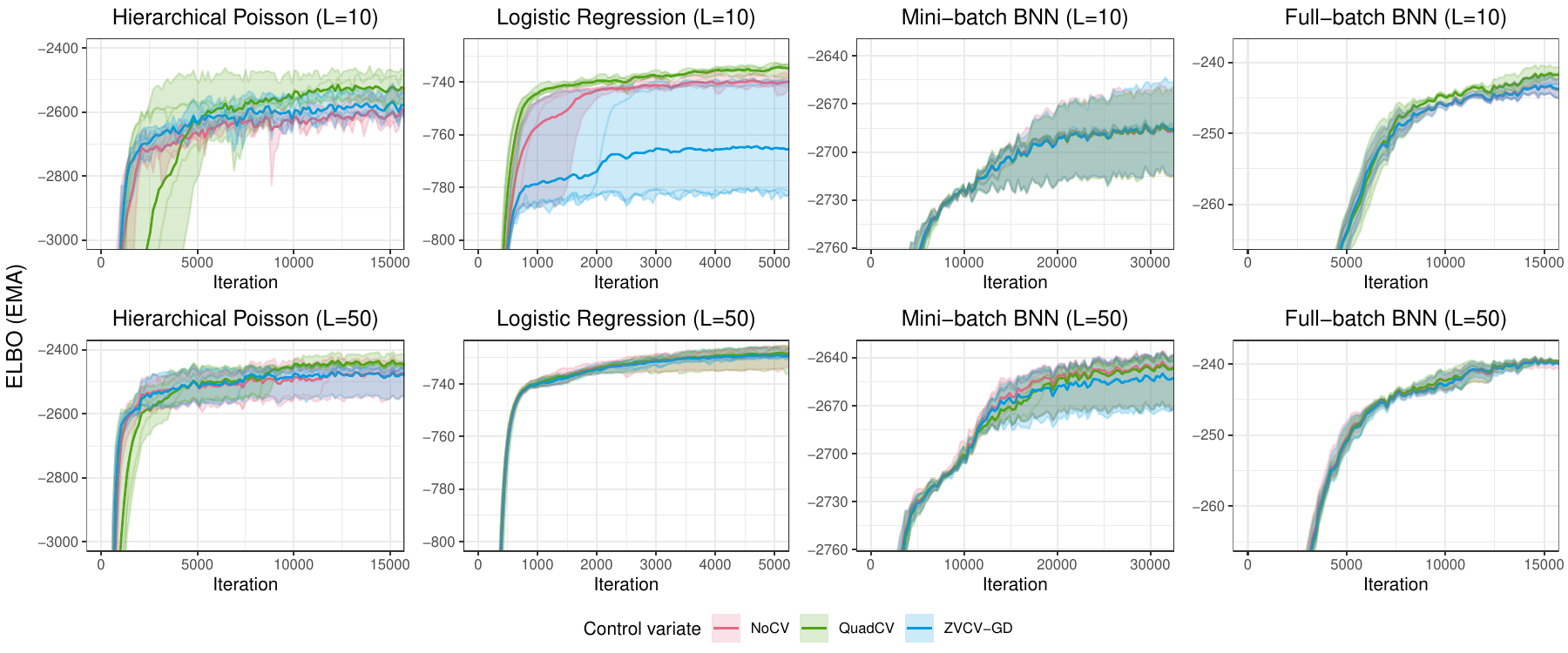}
    \caption{Real NVP with 10 (top) and 50 (bottom) gradient samples.}
    \label{fig:mean-elbo-iter-nvp}
  \end{subfigure}
  \caption{ELBO is plotted against the number of gradient descent steps for different
    numbers of gradient samples $\nmc$ and two families of $\qlbda$. The bold lines
    represent the mean of ELBO values recorded at the same iteration across five
    repetitions. The shaded area illustrates the range of ELBO values across five
    repetitions. The ELBO values are smoothed using an exponential moving average. The
    trajectories of ZVCV-GD and NoCV are nearly identical in both full-batch and
    mini-batch BNN when $L=10$. A higher ELBO indicates better performance. See
    Figure~\ref{fig:elbo-iter} for plots where the bold lines represent the median
    ELBO.}
  \label{fig:mean-elbo-iter}
\end{figure}

\begin{figure}
  \centering
  \includegraphics[width=\linewidth]{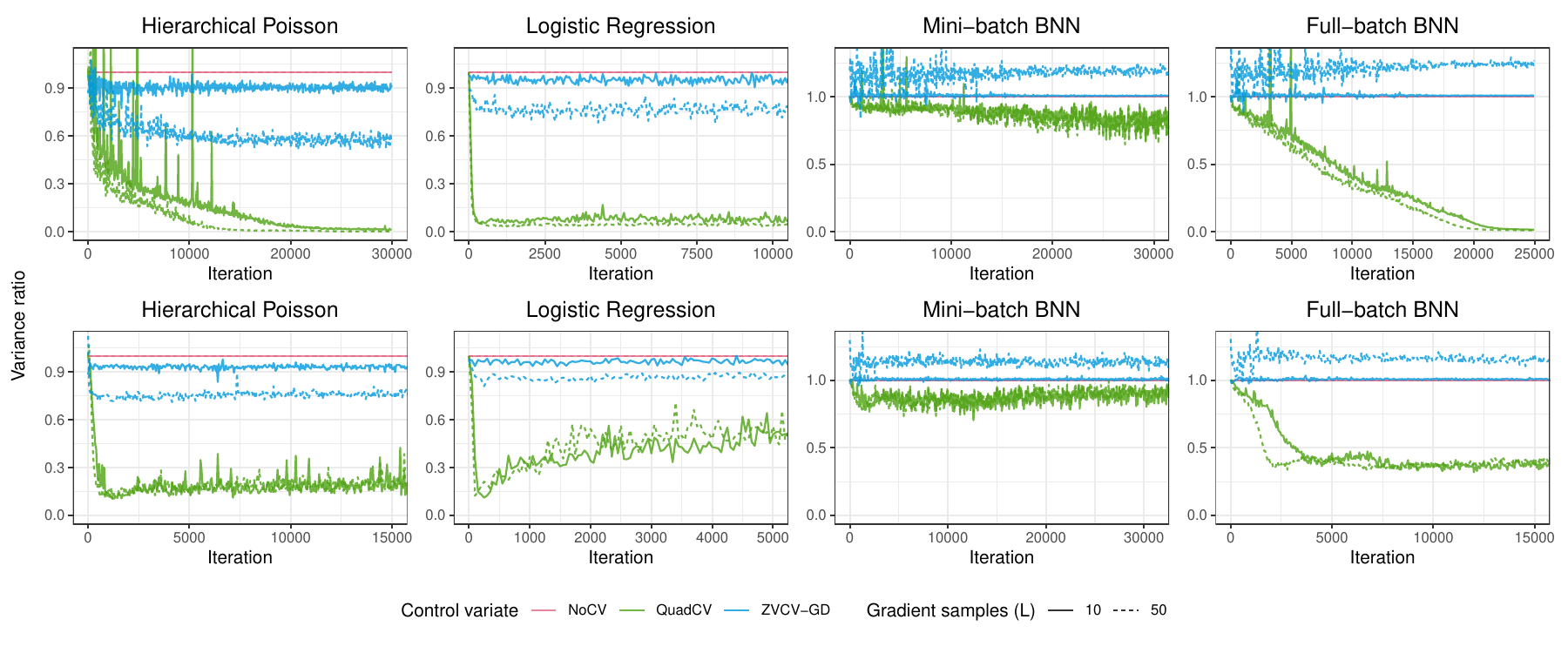}
  \caption{We present the variance ratio $\Var[\hat \gelbocv] / \Var[\hat \gelbo]$,
    where $\hat \gelbo$ is NoCV and $\hat \gelbocv$ is either ZVCV-GD or QuadCV, at each
    iteration. We show only the mean variance ratios recorded at the same iteration
    across five repetitions, omitting the individual variance ratios from each
    repetition to prevent clutter in the plots. The ratios from mean-field Gaussian and
    real NVP are shown in top and bottom rows respectively. Note that NoCV (in red) is
    always 1 by definition. We see that ZVCV-GD (in blue) struggles to reduce variance
    in the BNN models. There is also a significant overlap in QuadCV between $\nmc = 10$
    (solid green) and $\nmc = 50$ (dotted green). A lower ratio indicates better
    performance. See Figure~\ref{fig:vr} for plots where the bold lines represent the
    median variance ratios.}
  \label{fig:mean-vr}
\end{figure}

\begin{figure}
  \centering
  \begin{subfigure}[t]{\linewidth}
    \centering
    \includegraphics[width=\linewidth]{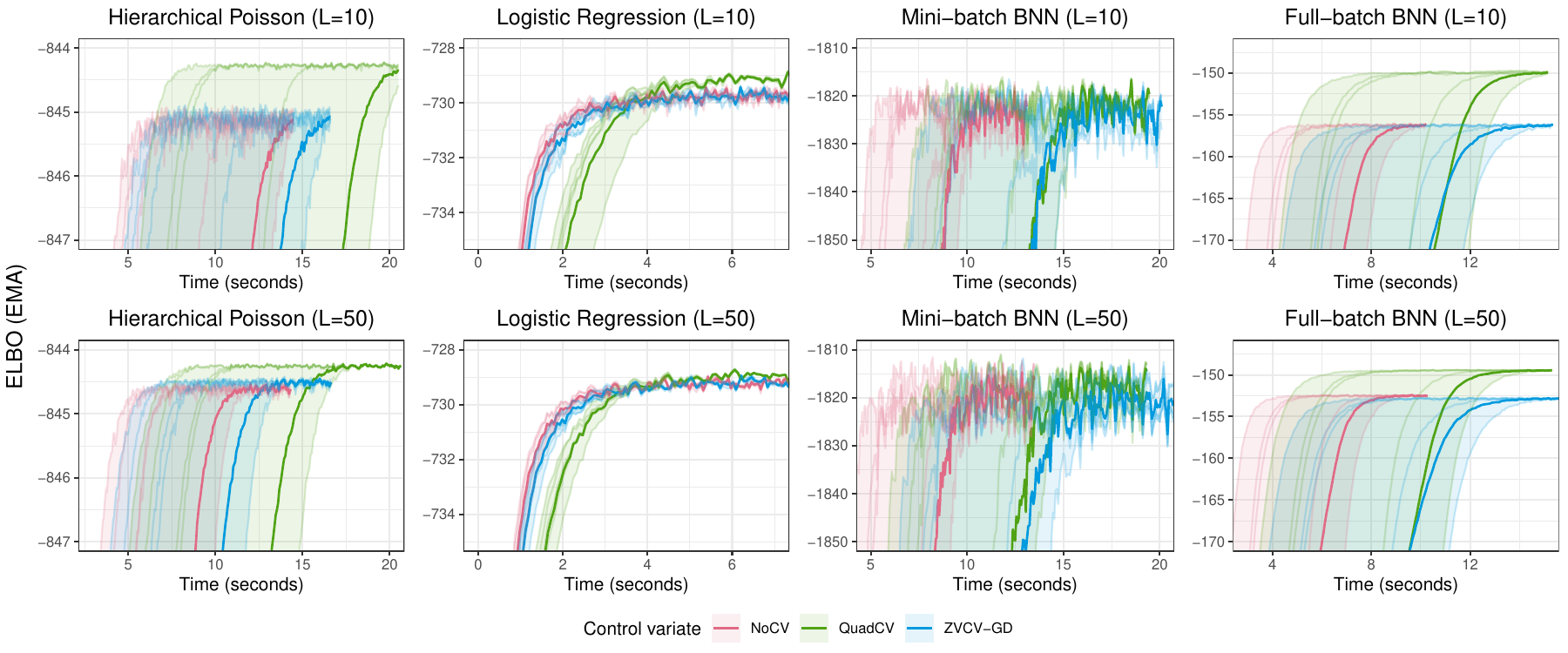}
    \caption{Mean-field Gaussian with 10 (top) and 50 (bottom) gradient samples.}
    \label{fig:mean-elbo-time-diag}
  \end{subfigure} \\
  \vspace{5mm}
  \begin{subfigure}[t]{\linewidth}
    \centering
    \includegraphics[width=\linewidth]{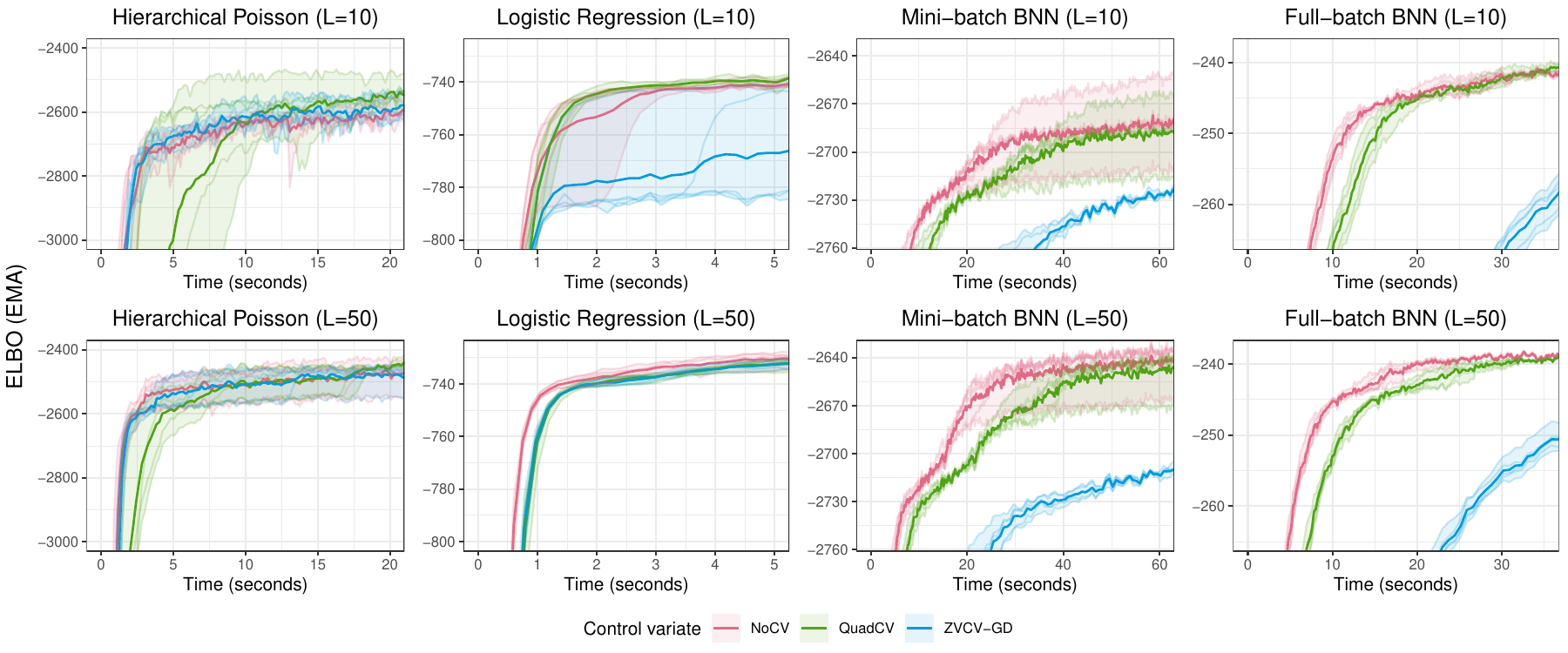}
    \caption{Real NVP with 10 (top) and 50 (bottom) gradient samples.}
    \label{fig:mean-elbo-time-nvp}
  \end{subfigure}
  \caption{ELBO is plotted against wall-clock time for different numbers of gradient
    samples $\nmc$ and two families of $\qlbda$. The bold lines represent the mean of
    ELBO values recorded at the same iteration across five repetitions. The shaded area
    illustrates the range of ELBO values across five repetitions. The ELBO values are
    smoothed using an exponential moving average. A higher ELBO indicates better
    performance. See Figure~\ref{fig:elbo-time} for plots where the bold lines represent
    the median ELBO.}
  \label{fig:mean-elbo-time}
\end{figure}

\begin{figure}
  \centering
  \begin{subfigure}[t]{\linewidth}
    \centering
    \includegraphics[width=\linewidth]{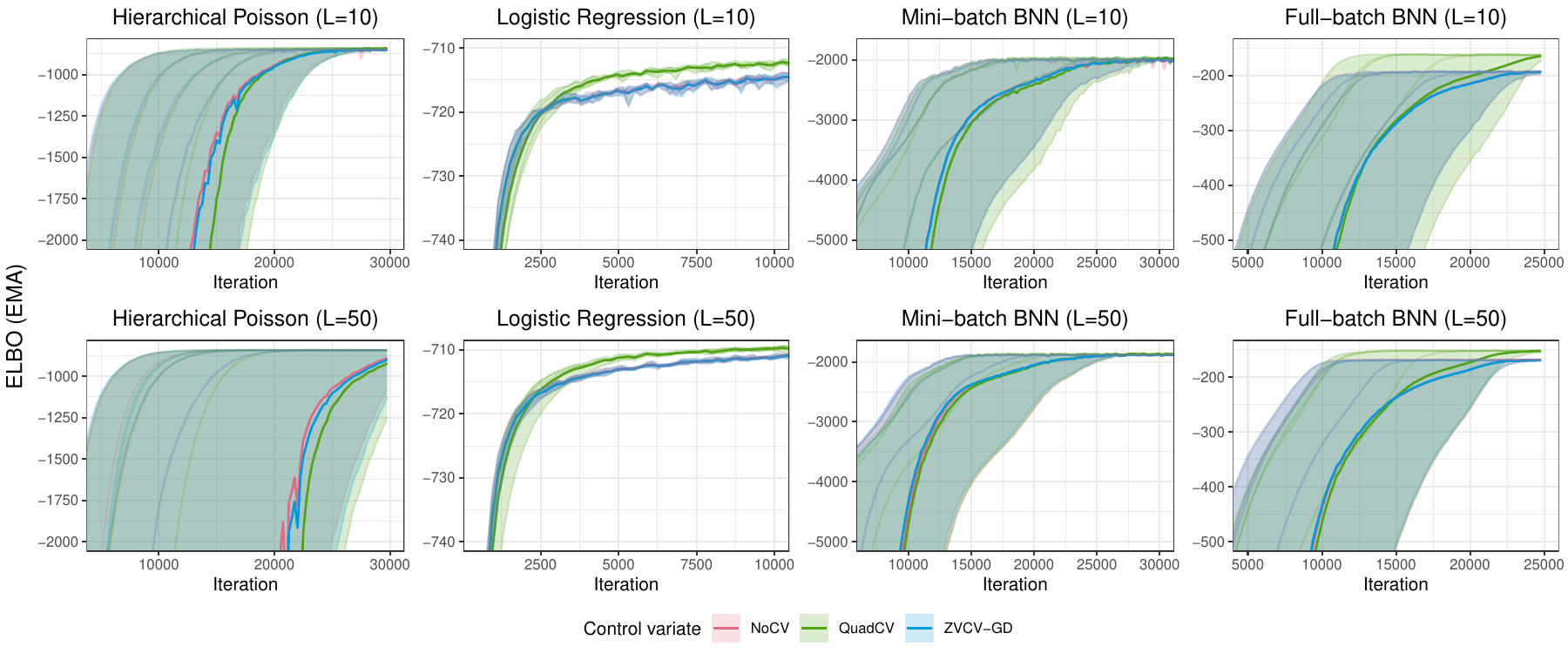}
    \caption{ELBO versus iteration counts.}
    \label{fig:mean-elbo-iter-diaglr}
  \end{subfigure} \\
  \vspace{5mm}
  \begin{subfigure}[t]{\linewidth}
    \centering
    \includegraphics[width=\linewidth]{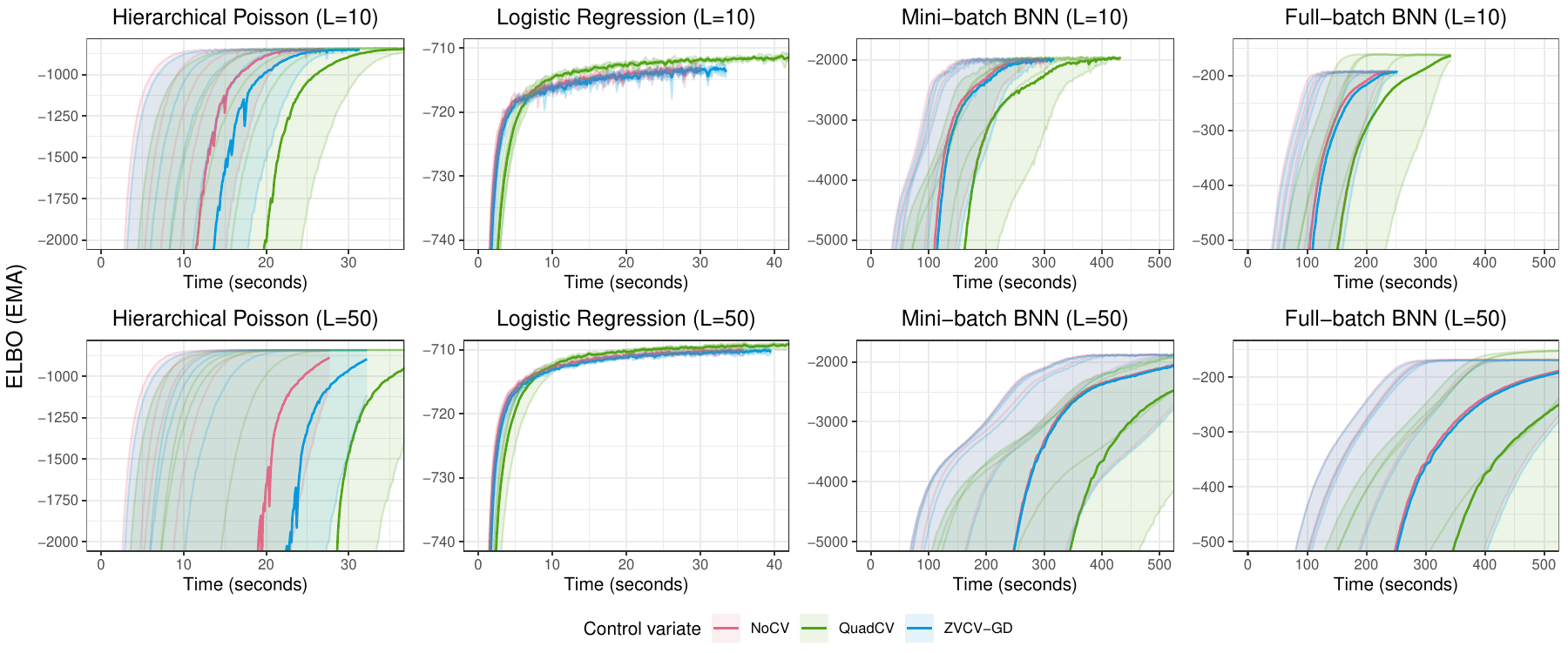}
    \caption{ELBO versus wall-clock time.}
    \label{fig:mean-elbo-time-diaglr}
  \end{subfigure}
  \caption{ ELBO is plotted against gradient descent steps and wall-clock time for
    varying numbers of gradient samples $\nmc$ using rank-5 Gaussian. The bold lines
    represent the mean of ELBO values recorded at the same iteration across five
    repetitions. The ELBO values have been smoothed using an exponential moving
    average. A higher ELBO indicates better performance. See Figure~\ref{fig:elbo-diaglr}
    for plots where the bold lines represent the median ELBO.}
  \label{fig:mean-elbo-diaglr}
\end{figure}

\begin{figure}
  \centering
  \begin{subfigure}[t]{\linewidth}
    \centering
    \includegraphics[width=\linewidth]{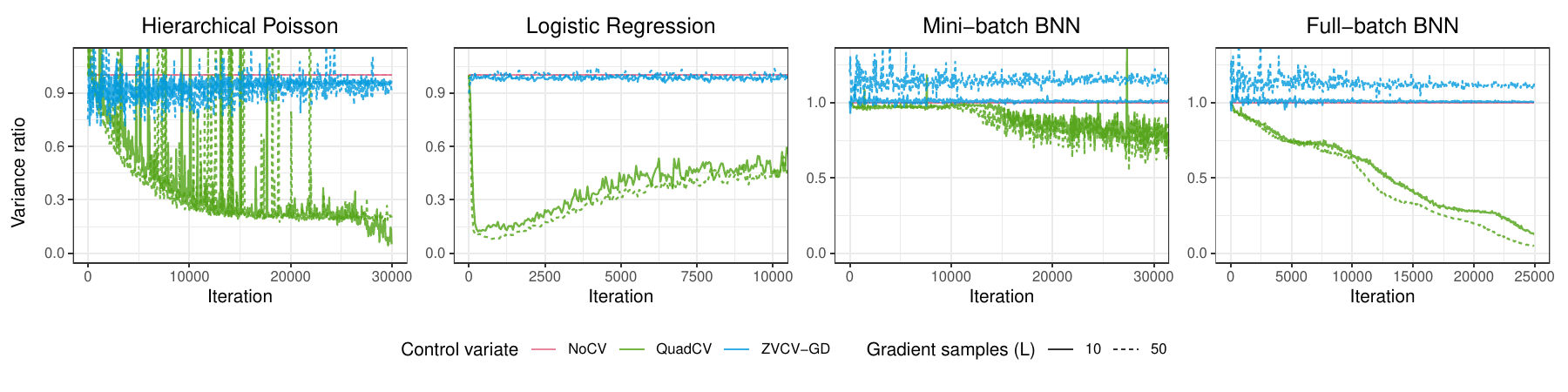}
    \caption{Variance ratio}
  \end{subfigure}
  \caption{We present the variance ratio $\Var[\hat \gelbocv] / \Var[\hat \gelbo]$ of
    rank-5 Gaussian, where $\hat \gelbo$ is NoCV and $\hat \gelbocv$ is either ZVCV-GD
    or QuadCV, at each iteration. We show only the mean variance ratios recorded at the
    same iteration across five repetitions, omitting the individual variance ratios from
    each repetition to prevent clutter in the plots. Note that NoCV (in red) is always 1
    by definition. We see that ZVCV-GD (in blue) struggles to reduce variance in the BNN
    models. There is also some overlap between $\nmc = 10$ (solid green) and $\nmc = 50$
    (dotted green). A lower ratio indicates better performance. See
    Figure~\ref{fig:vr-diaglr} for plots where the bold lines represent the median
    variance ratios. }
  \label{fig:mean-vr-diaglr}
\end{figure}

\section{Individual runs of full-batch BNN with mean-field Gaussian}
\label{sec:different-initialisation}
We zoom in on a particular model and variational family from the experiments in the main
text. Our aim in this section is to look the trajectory according to each initialisation
separately to help visualise the impact of initialisation on convergence. Due to space
limitations, we have only included trajectories from full-batch BNN with mean-field
Gaussian. In addition to the ELBO reported in the main text, we also report the
downstream metric, log pointwise predictive density evaluated on a test set (test lppd),
which is popular in the VI literature. Mathematically, the test lppd is defined as
\begin{equation*}
  \sum_{x \in \mathcal{D}_{\operatorname{test}}} \log \left( |\mathcal{Z}|^{-1} \sum_{z \in \mathcal{Z}} p(x | z)\right).
\end{equation*}
Here, $\mathcal{D}_{\text{test}}$ represents a test set, $\mathcal{Z}$ is a set
of samples drawn from $\qlbda(z; \lambda)$, and $|\mathcal{Z}|$ indicates the
cardinality of $\mathcal{Z}$. We have set $|\mathcal{Z}| = 1000$ in our
experiments. The test lppd is also referred to as the test log-likelihood, test
log-predictive, or predictive log-likelihood in the literature \citeP{yao2019,
  deshpande2022}.

\begin{figure}[h!]
  \centering
  \begin{subfigure}[t]{\linewidth}
    \centering
    \includegraphics[width=\linewidth]{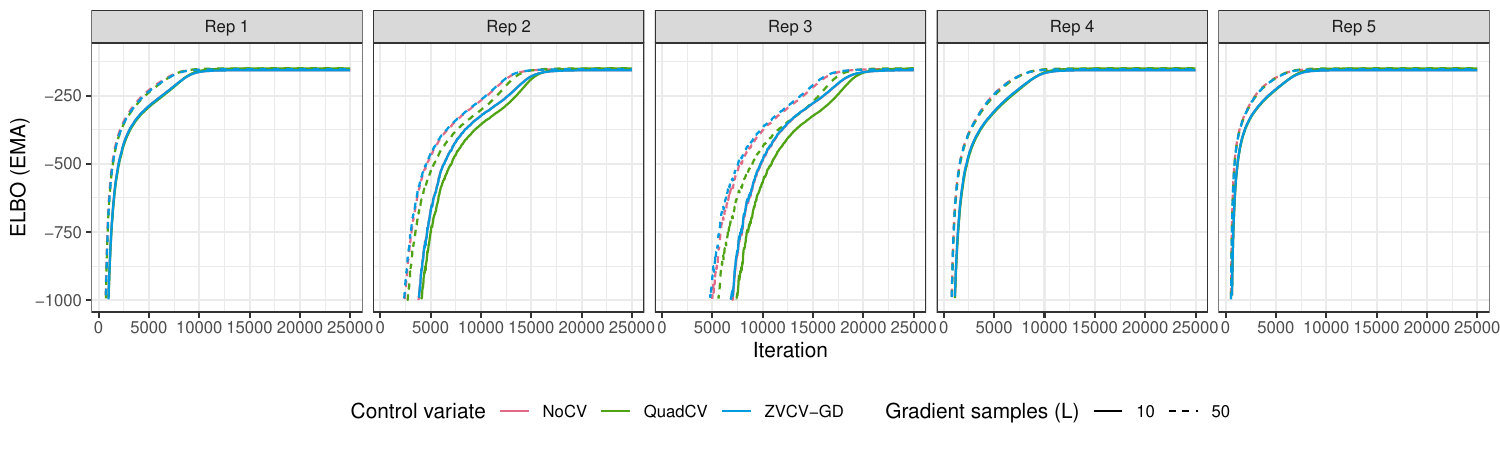}
    \caption{ELBO trajectories. This is a zoomed-out version of the last column of
      Figure~\ref{fig:elbo-iter-diag}. Higher values are preferred.}
    \label{fig:elbo-iter-5reps}
  \end{subfigure} \\
  \vspace{5mm}
  \begin{subfigure}[t]{\linewidth}
    \centering
    \includegraphics[width=\linewidth]{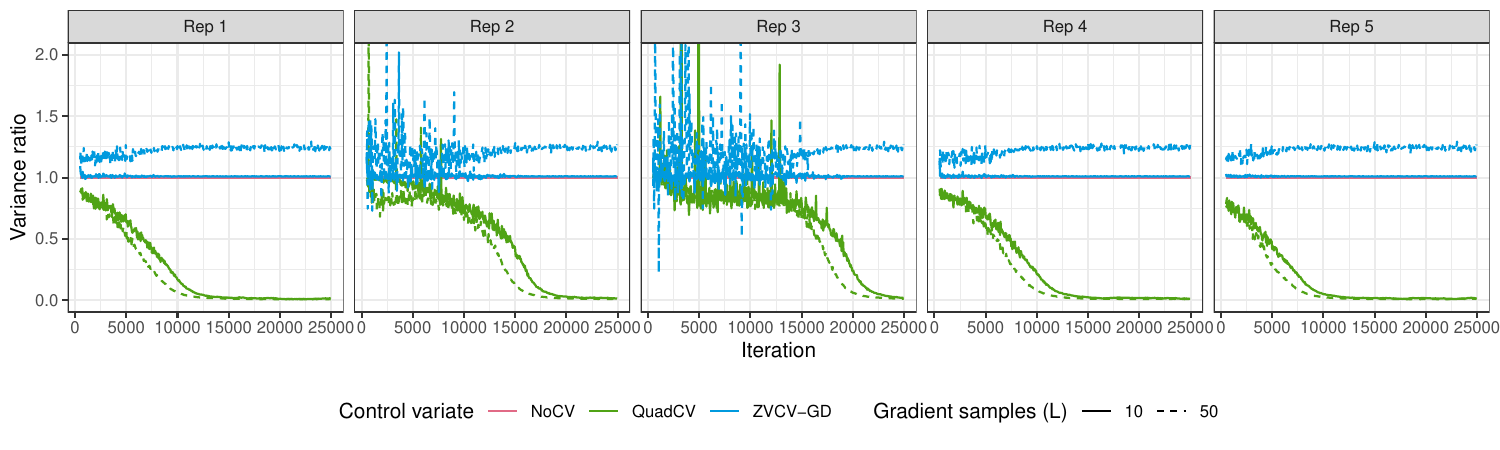}
    \caption{Variance ratios. A reading of 1 indicates no variance reduction. Lower
      values are preferable.}
    \label{fig:vr-5reps}
  \end{subfigure}
  \caption{The trajectories of ELBO and variance ratio for full-batch BNN with
    mean-field Gaussian are depicted over the course of iterations, with each of the
    five repetitions presented individually. By definition, the variance ratio of NoCV
    (red) is 1. Notably, there is a substantial overlap between NoCV (in red) and
    ZVCV-GD (in blue). In some cases, the distinctions between all three methods are
    hardly discernible. However, there is a relatively noticeable difference between
    $\nmc = 10$ and $\nmc = 50$.}
  \label{fig:elbo-vr-5reps}
\end{figure}

Figures~\ref{fig:elbo-iter-5reps} clearly show that trajectories vary substantially with
different initialisations. This is consistent with the high variability of ELBO
trajectories in Figure~\ref{fig:elbo-iter}.

In all cases, increasing $\nmc$, the number of gradient samples, effectively reduces the
variance of the gradient estimator from the outset of the optimisation process. This
stands in contrast to QuadCV, which only becomes effective after the quadratic
approximation $\tilde \dt$ in~\eqref{eq:quad-cv} has been adequately trained
(Figure~\ref{fig:vr-5reps}). Consequently, QuadCV performs poorly in the early and
middle stages of optimisation (as seen in Repetitions 2 and 3 in
Figure~\ref{fig:elbo-iter-5reps}).

\begin{figure}[h!]
  \centering
  \begin{subfigure}[t]{\linewidth}
    \centering
    \includegraphics[width=\linewidth]{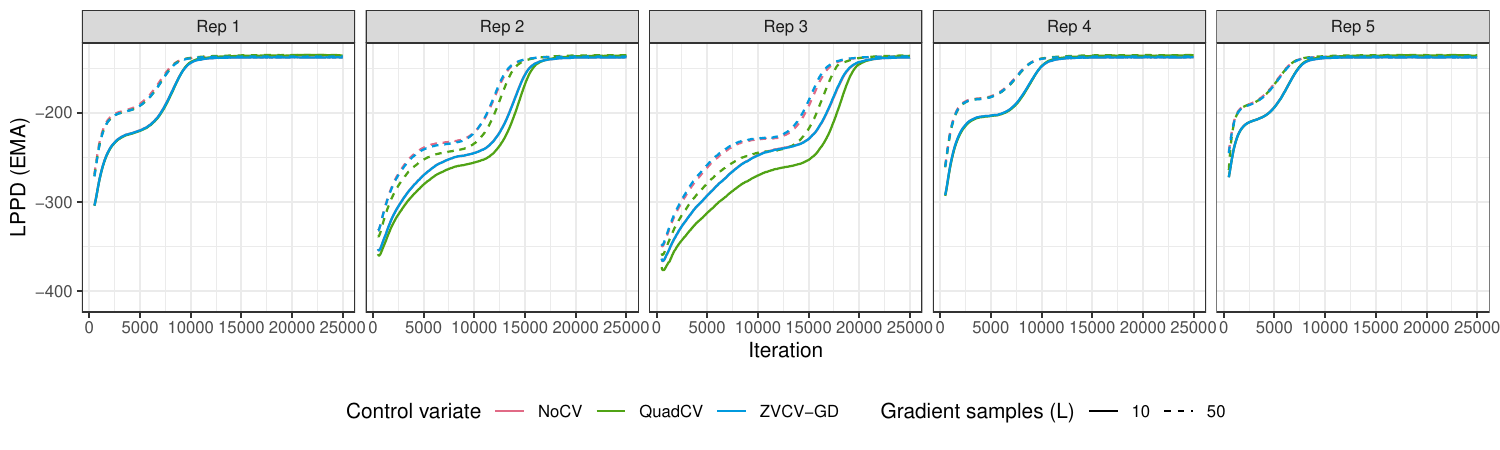}
    \caption{Test lppd trajectories.}
    \label{fig:lppd-iter-5reps}
  \end{subfigure} \\
  \vspace{5mm}
  \begin{subfigure}[t]{\linewidth}
    \centering
    \includegraphics[width=\linewidth]{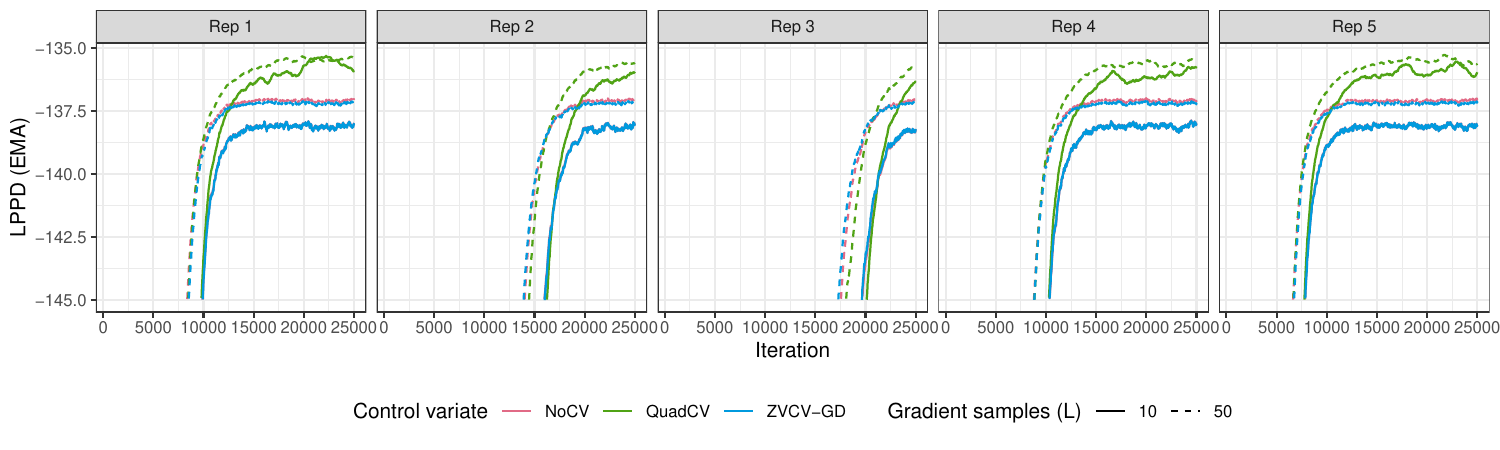}
    \caption{Test lppd trajectories, zooming in between
      $\operatorname{lppd} = (-145, -135)$.}
    \label{fig:lppd-iter-5reps-zoom}
  \end{subfigure}
  \caption{The trajectories of test lppd for full-batch BNN with mean-field Gaussian are
    depicted over the course of iterations, with each of the five repetitions presented
    individually. Notably, there is a substantial overlap between NoCV (in red) and
    ZVCV-GD (in blue). In some cases, the distinctions between all three methods are
    hardly discernible. However, there is a relatively noticeable difference between
    $\nmc = 10$ and $\nmc = 50$. Higher values are preferred.}
  \label{fig:lppd-5reps}
\end{figure}

Prior research on variance reduction in pathwise gradient estimators
\citeP{miller2017, geffner2018, geffner2020} often aims to push the boundaries
of attainable ELBO. Achieving this typically requires longer training periods. However,
we are of the opinion that the additional ELBO gained through this effort does not
warrant the extra computational cost incurred by implementing control variates. This is
particularly relevant given that improvements in downstream metrics, such as test lppd,
are marginal when compared to improvements achieved in the earlier stages of
optimisation (note the y-axis scale in Figure~\ref{fig:lppd-iter-5reps}
and~\ref{fig:lppd-iter-5reps-zoom}).

For instance, in Repetition 1, there is only a 3 nats improvement in test lppd (over a
test set of size 100), while substantial improvements are observed in the earlier
stages, often in the scale of hundreds. These 3 nats come at a cost of over 50\%
additional computation time compared to NoCV (as indicated in
Figure~\ref{fig:elbo-time-diag}). Furthermore, it is worth noting that an improvement in
ELBO does not invariably guarantee a substantial improvement in downstream statistics,
as evidenced in previous works, such as \citeT{yao2018a, yao2019, foong2020,
  masegosa2020, deshpande2022}.

\section{Comparison of ZVCV-GD with different hyperparameters}
\label{sec:zvcv-hyperparameters}
\begin{figure}[h!]
  \centering
  \begin{subfigure}[t]{\linewidth}
    \centering
    \includegraphics[width=\linewidth]{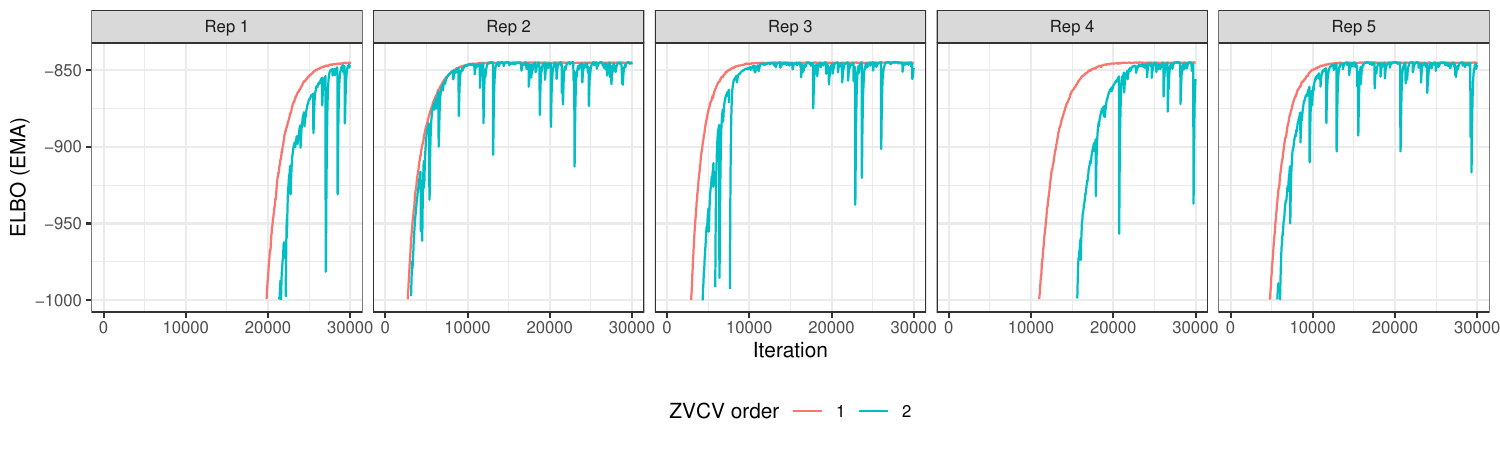}
    \caption{ELBO trajectories. Higher values are preferable.}
    \label{fig:elbo-zvcv-orders}
  \end{subfigure} \\
  \vspace{5mm}
  \begin{subfigure}[t]{\linewidth}
    \centering
    \includegraphics[width=\linewidth]{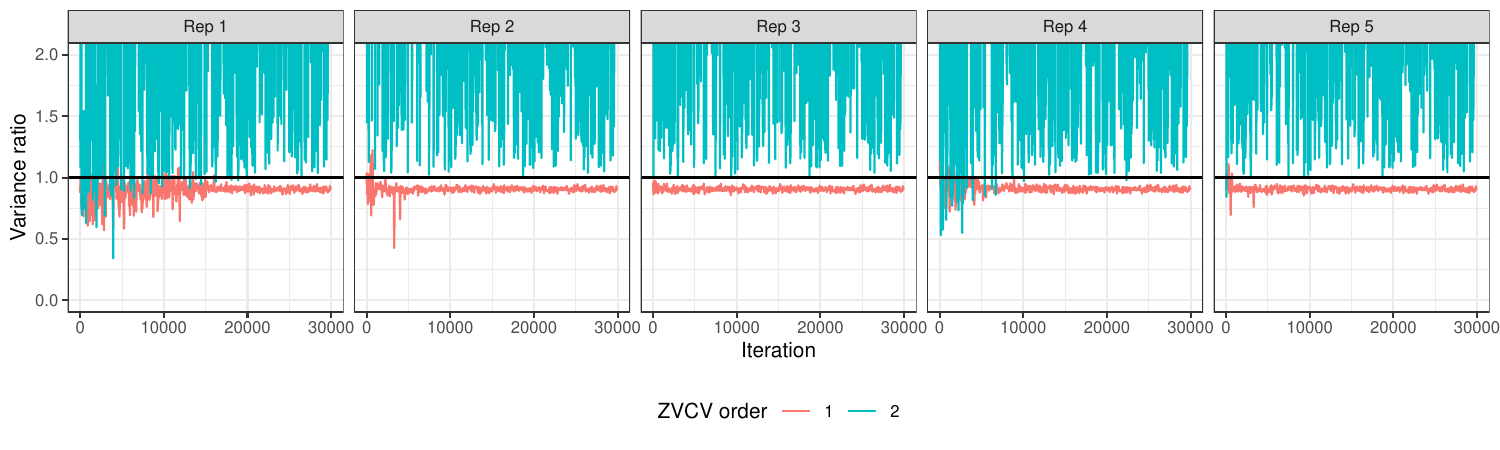}
    \caption{Variance ratios. A reading of 1 indicates no variance reduction. Lower
      values are preferable.}
    \label{fig:vr-zvcv-orders}
  \end{subfigure}
  \caption{ELBO trajectories and variance ratios for hierarchical Poisson models using
    mean-field Gaussian, ZVCV-GD with $\nmc = 10$, and first- and second-order
    ZVCV-GD both with 4 inner GD steps. The experiment was repeated five times, each time with different
    initialisations.}
  \label{fig:zvcv-orders}
\end{figure}

\begin{figure}[h!]
  \centering
  \begin{subfigure}[t]{\linewidth}
    \centering
    \includegraphics[width=\linewidth]{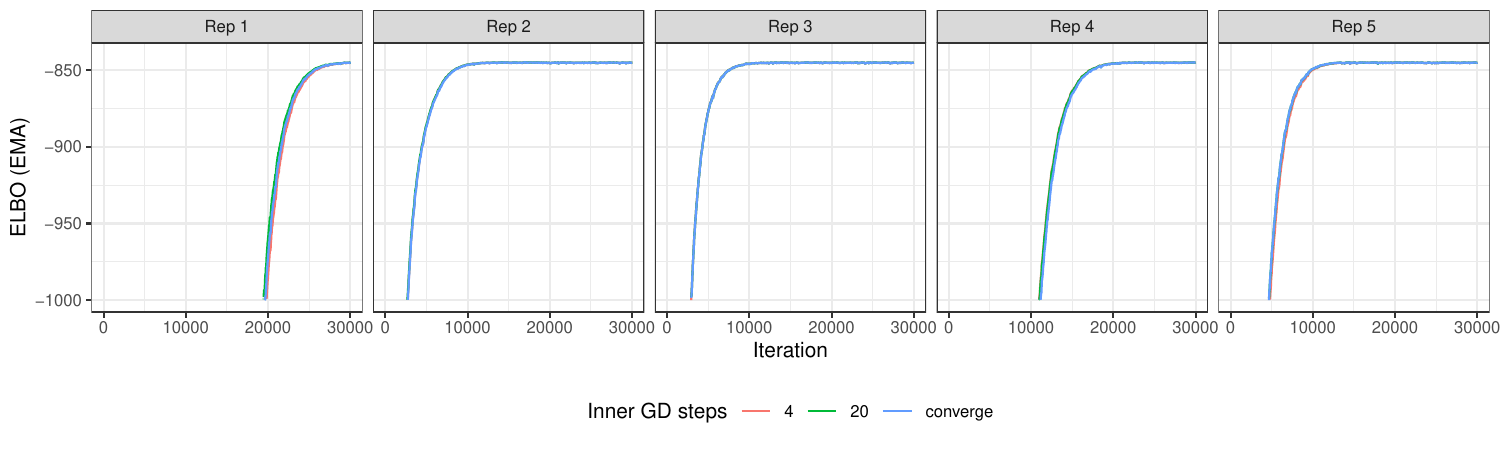}
    \caption{ELBO trajectories. Higher values are preferable.}
    \label{fig:elbo-zvcv-converge}
  \end{subfigure} \\
  \vspace{5mm}
  \begin{subfigure}[t]{\linewidth}
    \centering
    \includegraphics[width=\linewidth]{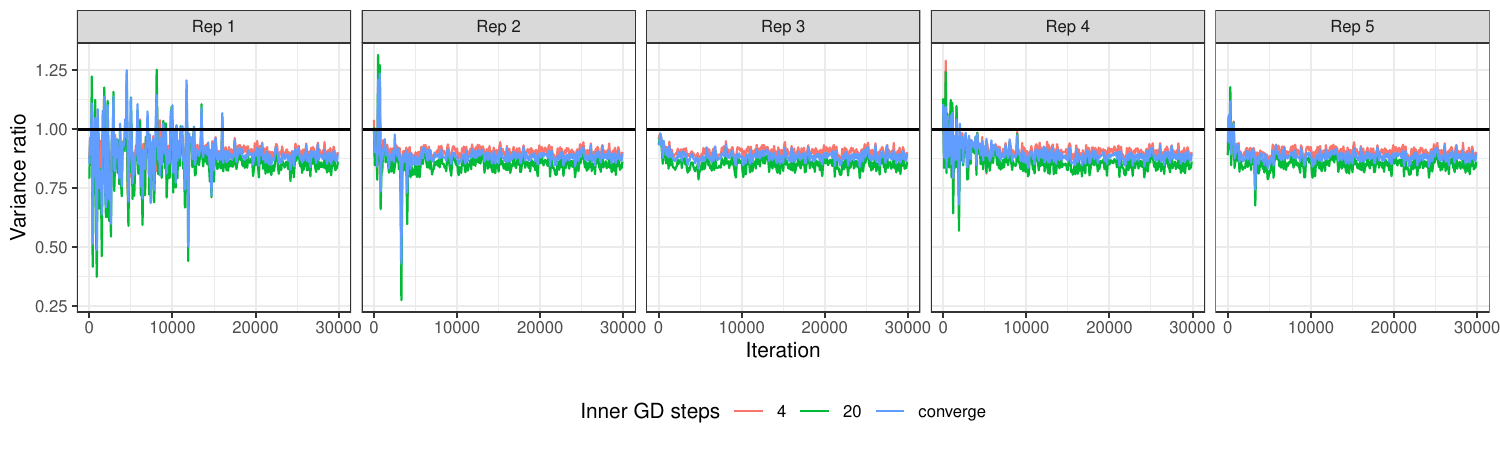}
    \caption{Variance ratios. A reading of 1 indicates no variance reduction. Lower
      values are preferable.}
    \label{fig:vr-zvcv-converge}
  \end{subfigure}
  \caption{ELBO trajectories and variance ratios for hierarchical Poisson models using
    mean-field Gaussian, (first-order) ZVCV-GD with $\nmc = 10$, running with different
    number of steps in the inner gradient descent. The experiment was repeated five
    times, each time with different initialisations. The ELBO trajectories for different
    GD steps are practically indistinguishable. The erratic variance ratio readings
    occur during the early optimisation stages in the low ELBO region, where gradient
    magnitudes are substantial.}
  \label{fig:zvcv-converge}
\end{figure}

We conducted experiments with ZVCV-GD that explore various hyperparameter settings,
running with both first- and second-order polynomials (Figure~\ref{fig:zvcv-orders}),
and testing different number of steps in the inner gradient descent loop
(Figure~\ref{fig:zvcv-converge}). We focus on the hierarchical Poisson model using a
mean-field Gaussian and setting $\nmc = 10$. We repeated the experiment five times, each
time with different initialisations. The red trajectories in
Figure~\ref{fig:zvcv-orders} and~\ref{fig:zvcv-converge} correspond to the default
settings of ZVCV-GD as specified in Section~\ref{sec:experiments}.

Figure~\ref{fig:vr-zvcv-orders} reveals that second-order ZVCV-GD did not effectively
reduce variance in the gradient estimator; instead, it introduced additional noise into
the estimator. This detrimental impact is also evident in the ELBO trajectories, as
shown in Figure~\ref{fig:elbo-zvcv-orders}. In light of these findings, we concluded
that the simpler first-order ZVCV-GD is preferable over the second-order variant.

In Figure~\ref{fig:zvcv-converge}, we present the ELBO trajectories and variance ratios
obtained by running the inner gradient descent (GD) of ZVCV-GD with three different
settings: 4 steps, 20 steps, and `until convergence'. Here, `convergence' is defined as
the point at which the residual of the inner least squares problem
in~\eqref{eq:ols-residual} no longer decreases substantially.

We observe that running the inner GD until convergence does not necessarily yield the
greatest variance reduction, as illustrated in Figure~\ref{fig:vr-zvcv-converge}. This
phenomenon can be attributed to overfitting the linear regression
in~\eqref{eq:ols-residual}, where the number of rows in $\CV$ is considerably smaller
than the number of columns. On the other hand, iterating the inner GD 20 times achieves
a more substantial variance reduction compared to the default 4 steps.

However, it is worth highlighting that there is no discernible impact on the ELBO
trajectories when varying the number of GD steps, as demonstrated in
Figure~\ref{fig:elbo-zvcv-converge}.

The optimal number of steps is not always evident without experimentation. Hence, we
typically opt for 4 steps to balance computational efficiency and the risk of
over-optimizing the inner GD process.

\end{document}